\documentclass[a4paper, 12pt]{article}

\usepackage{sbc-template}
\usepackage{booktabs}
\usepackage{tabularx}
\usepackage[table]{xcolor}
\usepackage{enumitem}
\usepackage{graphicx}
\usepackage[cmex10]{amsmath}
\usepackage{multirow}
\usepackage{color}
\usepackage{amssymb}
\usepackage{import}
\usepackage{flushend}
\usepackage{array}
\usepackage{url}
\usepackage{subcaption}
\usepackage{float}
\usepackage[numbers]{natbib}
\title{ForestEyes Project: Conception, Enhancements, \\and Challenges\footnote{
Published at Elsevier Future Generation Computer System (FGCS), Volume 124, November 2021, Pages 422-435 [https://doi.org/10.1016/j.future.2021.06.002]
}}

\author{Fernanda B. J. R. Dallaqua, Álvaro Luiz Fazenda, Fabio A. Faria }

\address{Instituto de Ciência e Tecnologia -- Universidade Federal de São Paulo
  (ICT-UNIFESP)\\
  Avenida Cesare Mansueto Giulio Lattes, n\textsuperscript{\underline{o}} 1201 -- Eugênio de Mello -- SP -- Brasil
\email{\{fernanda.dallaqua,alvaro.fazenda,ffaria\}@unifesp.br}
}

\begin{document} 

\maketitle

\begin{abstract}

Rainforests play an important role in the global ecosystem. However, significant regions of them are facing deforestation and degradation due to several reasons.
Diverse government and private initiatives were created to monitor and alert for deforestation increases from remote sensing images, using different ways to deal with the notable amount of generated data.
Citizen Science projects can also be used to reach the same goal.
Citizen Science consists of scientific research involving nonprofessional volunteers for analyzing, collecting data, and using their computational resources to outcome advancements in science and to increase the public's understanding of problems in specific knowledge areas such as astronomy, chemistry, mathematics, and physics. 
In this sense, this work presents a Citizen Science project called ForestEyes, which uses volunteer's answers through the analysis and classification of remote sensing images to monitor deforestation regions in rainforests.
To evaluate the quality of those answers, different campaigns/workflows were launched using remote sensing images from Brazilian Legal Amazon and their results were compared to an official groundtruth from the Amazon Deforestation Monitoring Project PRODES. 
In this work, the first two workflows that enclose the State of Rondônia in the years 2013 and 2016 received more than $35,000$ answers from $383$ volunteers in the $2,050$ created tasks in only two and a half weeks after their launch.
For the other four workflows, even enclosing the same area (Rondônia) and different setups (e.g., image segmentation method, image resolution, and detection target), they received $51,035$ volunteers' answers gathered from $281$ volunteers in $3,358$ tasks.
In the performed experiments, it was possible to observe that the volunteers achieved satisfactory overall accuracy in the classification of forestation and non-forestation areas using the ForestEyes project. Furthermore, considering an efficient segmentation and a better image resolution, they can achieve outstanding effectiveness results in the classification task of recent deforestation images. Therefore, these results show that Citizen Science might be a powerful tool in monitoring deforestation regions in rainforests as well as in obtaining high-quality labeled data.
\end{abstract}

%\begin{keyword}
%Citizen Science\sep Deforestation area detection\sep Rainforest\sep Tropical forest\sep Volunteered Thinking
%\end{keyword}

%\%end{frontmatter}

%\linenumbers
\clearpage
\section{Introduction}

Rainforests have an important role in the global ecosystem once they have a great diversity of fauna and flora, regulate the climate and rainfall, absorb large amounts of carbon dioxide, and are indigenous housing~\cite{FAO}.
Unfortunately, thousands of square kilometers of rainforests are deforested or degraded daily due to livestock, agriculture, urban area expansion, wood extraction, and forest fires~\cite{AFW14}.

According to one of the most well-known and successful rainforests monitoring programs - PRODES~\cite{AFW14,PRODES,metodprodes2019} - in the period between August/$2018$ and July/$2019$ the Brazilian Legal Amazon's deforestation reached $10,129km^2$, an increase of $34$\% in comparison with the previous period (August/$2017$ to July/$2018$) ~\cite{prodes2019}. These high rates of deforestation can have irreversible and catastrophic consequences as biodiversity loss, climate change, desertification, water shortage, increase of diseases, and even emergence of pandemics~\cite{martin2015edge,amazonscience2018,coura2010chagas,afelt2018bats}.

As the conservation of rainforests is urgently needed, monitoring programs were created by government agencies and non-profit institutions. These programs use remote sensing images, image processing, machine learning techniques, and specialists' photointerpretation to analyze, identify, and quantify forest cover changes~\cite{AFW14}.

The shortage of skilled labor and the high amount of data to be analyzed is a significant challenge for information and communication technology (ICT)~\cite{SOARES2010}. A possible solution to overcome this problem is to use Citizen Science wherein non-specialized volunteers collect, analyze, and classify data to solve several technical and scientific problems~\cite{GREY2009,SILVERTOWN2009}.

CS projects of a wide variety of research fields have drawn the attention of well-known scientific magazines such as Nature~\cite{science2018,gura2013citizen} and Science~\cite{guerrini2018citizen}. Recently, the U.S. \footnote{\url{https://www.citizenscience.gov/}} and Australian \footnote{\url{https://citizenscience.org.au/}} 
governments, and the European Union\footnote{\url{https://www.ecsite.eu/activities-and-services/projects/eu-citizenscience}}, have launched official programs to catalog and support CS projects~\cite{bonney2016theory}.

CS can be a valuable source of data for the Earth Observation field, which includes deforestation monitoring~\cite{FRITZ}. For CS projects ForestWatchers~\cite{AFW14}, EarthWatchers~\cite{schepaschenko2019recent}, and Geo-Wiki~\cite{fritz2012geo} the volunteers analyze and classify remote sensing images. These classifications generate deforestation maps or alerts. For the ForestWatcher project, the volunteers collect data \textit{in situ} to confirm deforestation alerts issued by Global Forest Watch~\cite{forestwatcher}.

In April $2019$ the ForestEyes Project\footnote{\url{https://www.zooniverse.org/projects/dallaqua/foresteyes}} was launched~\cite{foresteyes2019}. This CS project is hosted by the Zooniverse.org platform~\cite{ZOONIVERSE2013} and aims to ally CS with machine learning techniques to monitor rainforests' deforestation. The volunteers analyze and classify remote sensing images. These classifications will be used to train machine learning techniques that will classify new remote sensing images. These new data can provide complementary information to official monitoring programs such as PRODES or even generate information for areas where such programs don't exist.

%The feasibility of the ForestEyes project depends on the quality of the data generated by the volunteers. The volunteers analyzed and classified Landsat-8's~\cite{lauer1997landsat} remote sensing images from an area of BLA specifically of the State of Rondônia, and PRODES-based data was used as validation of the volunteers' contributions. To ensure the comparison between volunteers' and PRODES' classifications, the volunteers analyzed the same Landsat-8 images that were used by PRODES.  

The feasibility of the ForestEyes project depends on the quality of the data generated by the volunteers. The volunteers analyzed and classified Landsat-8's~\cite{lauer1997landsat} remote sensing images from an area of BLA, specifically of the State of Rondônia, wherein PRODES-based data was used as validation of the volunteers' contributions. It is worth mentioning that to ensure this validation, the volunteers analyzed the same Landsat-8 images used by PRODES.

This paper describes in detail all of the steps taken to validate the generated data, including the creation of the tasks in the Zooniverse.org platform, the remote sensing image processing techniques, the launch of the CS's campaigns, the user layout continuous enhancements, and the evaluation of the volunteers' contributions. 
The remainder of this work is organized as follows. Section~\ref{fundTeor} presents background about deforestation monitoring and CS. 
Section~\ref{sec:ForestEyes} presents the proposed ForestEyes Project. 
Section~\ref{results} shows the methodologies and some results for the CS's campaigns, whereas Section~\ref{sec:comparison} presents the comparison between the campaigns and analyses of the volunteers' contributions. Section~\ref{conclusao} presents the conclusions, and Section~\ref{future} defines the future steps for ForestEyes Project. 

\section{Background}
\label{fundTeor}
This section presents background about deforestation detection/monitoring and CS projects.
%, and a brief explanation about ForestWatchers' Correct Classification~\cite{AFW14}.%, the inspiration of {ForestEyes}.

\subsection{Deforestation Detection and Monitoring}
\label{sec:def}

The monitoring of rainforests is necessary because of their great biodiversity. Besides the well-known Amazonian Forest, other rainforests also suffer from indiscriminate and progressive deforestation, where over $74$ million hectares have been lost or heavily degraded from $2000$ to $2012$~\cite{AFW14,Hansen850}. For BLA, there are monitoring programs - governmental and from non-profit institutions - for deforestation measuring and preventing as PRODES and DETER~\cite{DETER} from INPE, MapBiomas~\cite{souza2017mapbiomas} (multi-institutional initiative), SAD from Imazon~\cite{IMAZON} and GLAD  from Global Forest Watch~\cite{GLAD}. 

\subsubsection{PRODES}
Amazon Deforestation Monitoring Project (PRODES)~\cite{AFW14,metodprodes2019} was developed by INPE in $1988$ to carry annual deforestation surveys in BLA. Trained specialists classify remote sensing images through photointerpretation. These remote sensing images are usually from Landsat satellites, but other satellites as SENTINEL-2 and CBERS-4 can also be used. By $2013$, Landsat-8 was launched, which has a revisit time of $16$ days, $30$ meters of spatial resolution, and $11$ spectral bands.

\begin{figure*}[!ht]
\centering
\begin{tabular}{c}
\includegraphics[height=0.53\textwidth,keepaspectratio=true]{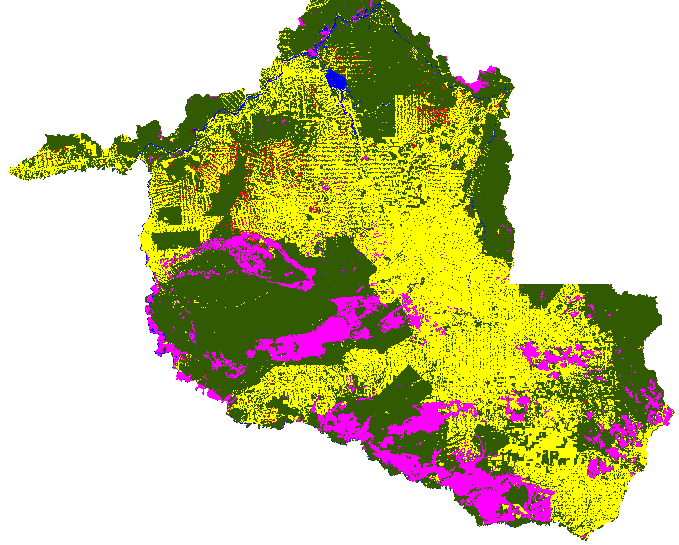} \\
(a) PRODES' mosaic of the State of Rondônia for the year of $2016$.\\
\includegraphics[height=0.25\textwidth,keepaspectratio=true]{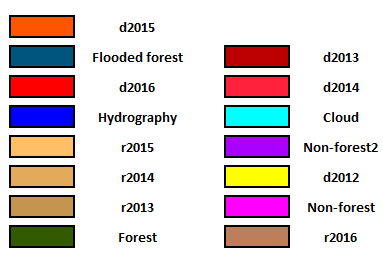} \\
 (b) Different classes existing on PRODES $2016$. \\
\end{tabular}
\caption{Rondônia's thematic image for the PRODES year of $2016$. Extracted from TerraBrasilis~\cite{deterrabrasilis}. In (a) Thematic image of Rondônia in the year $2016$ and in 
(b) Legend for the thematic image where "d" means deforestation of the given year and "r" is the residue. The residue is deforestation that was not detected but occurred in previous years~\cite{macielreasoning}.} 
\label{fig:imagensPRODES2016}
\end{figure*}

PRODES methodology is divided into three steps: (1) an image selection technique is performed, where cloud-free images collected near August 1st are enhanced to highlight the areas with complete removal of native forest (clear-cutting); (2) the specialists identify and delimit the deforestation polygons; and (3) the annual deforestation rate is calculated~\cite{metodprodes2019}. Both thematic images and quantitative data are publicly available. Figure~\ref{fig:imagensPRODES2016} presents the PRODES' mosaic for the State of Rondônia for the year $2016$.

As PRODES quantifies and spatially locates annual BLA deforestation increments, once an area is classified as deforestation, it is included in an exclusion mask that will be used in subsequent years. Therefore PRODES data don't infer on regenerated areas~\cite{PRODES3}. This exclusion mask also has regions of different vegetation (which PRODES classifies as non-forest) and hydrography areas.  

\subsubsection{DETER}

The monitoring program DETER was created to detect near-real-time deforestation as quick actions are needed to stop the beginning of a deforestation process. It began by using imagery from MODIS (Moderate Resolution Imaging Spectroradiometer)~\cite{MODIS} with a spatial resolution of $250$ meters and a revisit time of roughly $1.5$ days. An alert is sent to authorities when a deforestation process is identified. These alerts are also posted every month on the program's website. Nowadays, DETER also uses images from other satellites with different spatial resolutions to improve deforestation detection \cite{DETER}.

\subsubsection{TerraBrasilis}

Recently INPE deployed a new platform, called TerraBrasilis~\cite{deterrabrasilis}, which provides an interactive dashboard in which users can access and use spatial data generated by PRODES and DETER. Also, the data generated by INPE's monitoring programs are used by other programs as TerraClass \cite{almeida2016high,neves2017terraclass} and MapBiomas (Brazilian Annual Land Use and Land Cover Mapping Project) \cite{souza2017mapbiomas}, for instance.
The project TerraClass was launched in $2011$ through a partnership between INPE and EMBRAPA (Brazilian Agricultural Research Corporation) to produce maps showing land use and the cover of the deforested areas mapped by PRODES \cite{almeida2016high}.

\subsubsection{MapBiomas}

MapBiomas project began in $2015$, and it is a multi-institutional initiative to generate annual land use and land cover time series for all Brazilian territory, covering all six official Biomes \cite{souza2017mapbiomas}. 
Recently MapBiomas has launched an alert system from information generated by DETER, SAD, and GLAD to validate and refine data from daily high-resolution images with a spatial resolution of $3m$, creating surveys to be sent to different authorities.

\subsubsection{SAD and GLAD}

The monitoring programs SAD (Deforestation Alert System) and GLAD (Global Land Analysis \& Discovery) are held by the non-profit institutions Imazon (Institute for humanity and amazon environment) and GFW (Global Forest Watch), respectively. The first analyzes MODIS images with Normalized Difference Fraction Index (NDFI) and provides surveys every month on its website~\cite{IMAZON}. The second program uses Landsat images and decision trees to alert tree-cover loss not only in BLA but also in other areas with rainforests as Congo Basin and Southeast Asia~\cite{GLAD}.

\subsection{Citizen Science (CS)}

The collection, analysis, and classification of data for scientific research through volunteers' contributions (most of them are non-specialist) is known as CS~\cite{GREY2009,SILVERTOWN2009,science2018}.  The remotest idea of CS goes back more than $2000$ years, where people helped to monitor migratory locusts outbreaks destroying harvest in ancient China~\cite{science2018}.

The pioneer project called Christmas Bird Count~\cite{SILVERTOWN2009} created by ornithologist Frank Chapman in $1900$ proposes carrying out a census of the bird population as an alternative to hunting birds on Christmas. Currently, the project is held every year by the National Audubon Society. Besides ecology, CS nowadays is being used in a great variety of research fields as astronomy~\cite{PLANETHUNTERS,STARDUST}, chemistry~\cite{FOLDIT}, mathematics~\cite{POLYMATH}, and physics~\cite{ATLAS}. 

Advances and worldwide growing popularity for ICT resources remotely distributed (e.g., computers, tablets, and mobile phones) helped to increase the dissemination of CS projects \cite{SILVERTOWN2009}. CS can also be referred to as Citizen Cyberscience, which can be divided into three subcategories: Volunteered Computing, Volunteered Thinking, and Participatory Sensing~\cite{HAKLAY2013}.

\begin{itemize}
    \item \textbf{Volunteered Computing}: a distributed processing power is provided by several volunteers donating their computational resources. The first and best-known example of Volunteered Computing project is SETI@home \cite{SETI}, launched in May $1999$ by the UCLA/Berkeley. To participate, the volunteers execute a free program that downloads and analyzes radio telescope data, searching for extraterrestrial intelligence. LHC@home \cite{LHC2012} and climateprediction.net~\cite{CLIMATE2002} are other examples of projects. 
    
    \item \textbf{Volunteered Thinking}: the volunteers use their cognitive skills to analyze data or to perform different kinds of tasks. Galaxy Zoo, launched in $2007$, is a well-known example of this kind of CS, requiring, for more than a decade, volunteers to classify images of galaxies according to patterns available~\cite{AFW14}. Within $24$ hours of launch, they received almost $70,000$ classifications in one hour. More than $50$ million classifications were received during its first year by more than $150,000$ people. With the success of Galaxy Zoo, it evolved into Zooniverse.org, an online platform that hosts a wide variety of Volunteered Thinking projects. Until $2013$, it collected more than $300$ million analyses from $1$ million volunteers~\cite{ZOONIVERSE2013}. FoldIt~\cite{FOLDIT}, Stardust@home~\cite{STARDUST}, and Planet Hunters~\cite{PLANETHUNTERS} are other examples of Volunteered Thinking projects. 
    
    \item \textbf{Participatory Sensing}:  the volunteers are responsible for gathering data or information, usually \textit{in situ}. The eBird project, released in $2002$ by Cornell Lab of Ornithology and National Audubon Society, aims to engage citizen scientists to report bird observations through standardized protocols, documenting bird distribution, habitat, and patterns~\cite{EBIRD2014,EBIRD2009}. Other examples of participatory sensing projects are CoralWatch~\cite{CORALWATCH} and NoiseTube~\cite{NOISETUBE}. 
\end{itemize}

The motivation to volunteer contributions to a CS project was addressed by several authors. Altruism, for example, was cited as a major factor for CS projects related to nature conservation~\cite{COHN2008,BRADFORD2004,KING1998}. For projects on different purposes, like Galaxy Zoo and FoldIt, the main reasons were research contribution and the interest in science~\cite{RADDICK2009,CURTIS}. Other important factors are the importance of online communities~\cite{RADDICK2009,HOLOHAN} and competitiveness~\cite{HOLOHAN} through a volunteers' ranking. For the eBird project, for example, a substantial increase in data gathered, collecting more observations in one-month than in two years' project, appears after some changes that included competition between volunteers~\cite{HOCHACHKA}.

%The importance of online communities~\cite{RADDICK2009,HOLOHAN}, and competitiveness~\cite{HOLOHAN} through a volunteers' ranking were also pointed out as important factors. For the eBird project, for example, a substantial increase in data gathered, collecting more observations in one-month than in two years' project, appears after some changes that included competition between volunteers~\cite{HOCHACHKA}.

Many CS projects have shown that citizen scientists can produce high-quality data, being as efficient as specialists~\cite{COHN2008,ARCANJO,GREY2009,KOSMALA,foresteyes2019}.
However, statistical data analyses and validation heuristics or algorithms are also important to discard bad or non-reliable data received from volunteers, ensuring good data quality. To send redundant tasks to multiple volunteers, allowing to extract the information that constitutes the majority or consensus between the volunteers, is widely used as a good data quality mechanism.

Calibration tasks containing already classified samples by specialists could also be used as a quality mechanism since they allow a volunteers and specialists comparison.
In Galaxy Zoo, for example, only registered volunteers can participate, assigning weights to individual users according to the volunteer's skill~\cite{RADDICK2009,LINTOTT,ARCANJO}.

To overcome the data quality challenge in CS projects,  \citet{De_Lellis_2019} proposes a method using volunteers' diversity to improve data accuracy by a Bayesian inference algorithm.
A survey, gathering information about the volunteer's education level and the motivation to CS, is used to group the volunteers into distinct classes.
This method shows an improvement in data accuracy compared to majority voting.

\subsection{ForestWatchers Project}
The ForestWatchers was a CS project to monitor the rainforests through the volunteer's collaboration connected to the Internet~\cite{ARCANJO,AFW14}. It was launched in $2012$ by researchers of the Institute of Science and Technology of Federal University of S\~{a}o Paulo (ICT/UNIFESP), the Laboratory for Computing and Applied Mathematics at the National Institute for Space Research (LAC/INPE), and the Citizen Cyberscience Centre (CCC).

The project was composed of three experimental applications: (1) Best-Tile; (2) Deforestation Classification; and (3) Correct Classification. The Correct Classification application used PRODES classification and remote sensing images from MODIS sensor to train an artificial neural network (ANN)~\cite{ANN}. New remote sensing images were analyzed by this ANN, creating segmented images in Forest (represented by green pixels) and Non-Forest (displayed as red pixels). Tiles with size $3\times3$ pixels - corresponding to an area of approximately $56$ hectares - with confidence below a given threshold were sent for volunteers' inspection where they agreed or disagreed with the ANN's classification~\cite{AFW14}.

Two campaigns were performed, each one with an area of BLA: one from the State of Rond\^{o}nia for the year of $2011$, with $72$ tasks; and the other was from Aw\'{a}-Guaj\'{a} Indigenous Reserve in $2014$, with $26$ tasks.

\section{The Proposed ForestEyes Project}
\label{sec:ForestEyes}

The ForestEyes Project~\cite{foresteyes2019}, a CS project inspired by ForestWatchers Project and launched in April/$2019$, aims to monitor deforestation areas, where volunteers classify remote sensing image's segments that can be used as training sets of classification algorithms. These classification algorithms are used to label new segments, providing complementary information to official monitoring programs or even generating data for areas where do not exist this kind of programs.

\begin{figure*}[h]
    \centering
    \includegraphics[width=\textwidth]{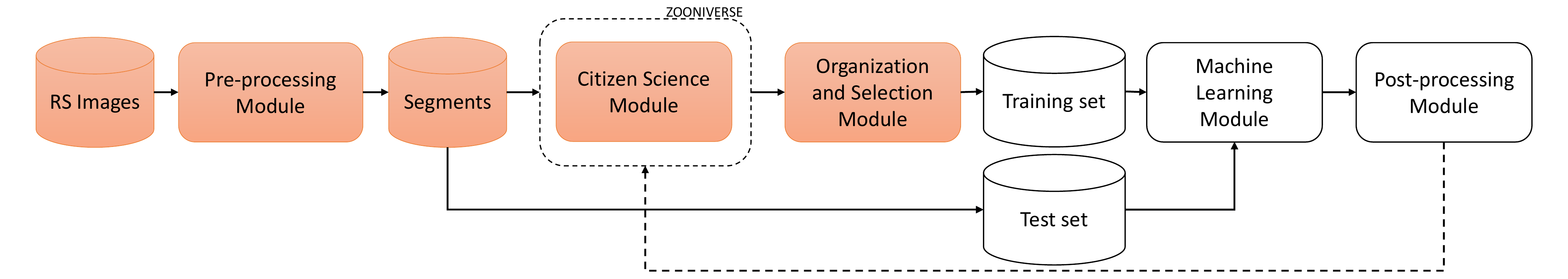}
    \caption{ForestEyes Project's schematic representation. The colored items are the modules discussed in this paper.}
    \label{fig:schemeFE2}
\end{figure*}

The project can be divided into five modules: (1) Pre-processing; (2) Citizen Science; (3)  Organization and Selection; (4) Machine Learning; (5) and Post-processing. A schematic representation of these modules can be seen in Figure~\ref{fig:schemeFE2} where the colored items are the ones discussed in this paper.

\subsection{Pre-processing Module}
\begin{figure*}%[!ht]
    \centering
    \includegraphics[width=1\textwidth]{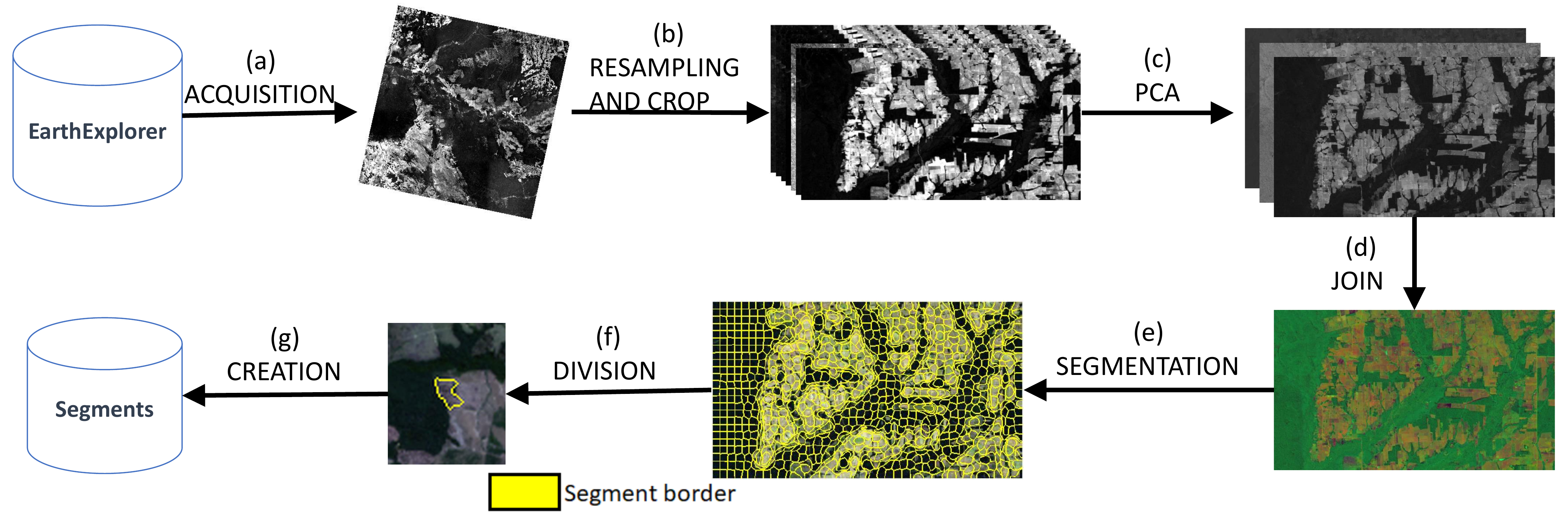}
    \caption{Schematic representation of the Pre-processing Module steps in the ForestEyes Project.}
    \label{fig:interface}
\end{figure*}

The Pre-processing Module is responsible for the acquisition, processing, and segmentation of the remote sensing images. Figure~\ref{fig:interface} shows the steps (a) to (g) of the Pre-processing Module. Firstly, in (a), the acquisition of the remote sensing images is performed. Currently, the images used by the project are from Landsat-8, the same ones that are analyzed by PRODES. The Landsat-8 scenes are freely collected at EarthExplorer platform\footnote{\url{https://earthexplorer.usgs.gov/}}. Landsat-8 has $11$ spectral bands but only $7$ bands - coastal aerosol or ultra-blue, blue, green, red, near-infrared or NIR, shortwave infrared 1 or SWIR 1, and shortwave infrared 2 or SWIR 2 - are acquired.  In (b),  the region of interest is cropped and resampled to standardize the remote sensing image according to the PRODES' mosaic.  

Generally, some segmentation algorithms have a $3$-band image (RGB) as input, and the acquired image is composed of $7$ bands. In (c), a dimensionality reduction technique (Principal Component Analysis - PCA) is applied to generate $3$ grey color images that are joined into one new $3$-band image, simulating RGB image in (d). This new image is used as input to a segmentation algorithm in (e). In this step, $3$ segmentation algorithms were used: (1) Simple Linear Iterative Clustering (SLIC)~\cite{SLIC}; (2) Image Foresting Transform-SLIC (IFT-SLIC)~\cite{alexandre2015ift}; and (3) MaskSLIC~\cite{irving2016maskslic}. 
In (f), each segment becomes a sample included in the database in (g). Some of the samples are sent for the Citizen Science Module, and the remaining samples compose the test set of the Machine Learning Module.

\subsection{Citizen Science  Module}

At the Citizen Science Module is performed the interaction with the volunteer, and The Zooniverse.org platform was chosen to host this module once it has a large and consolidated community. This platform allows the creation of one or more workflows,  defined as the sequence of tasks to be handled by the volunteers.
On the ForestEyes Project,  the volunteers analyze and classify the segments into three classes (Forest, Non-Forest, or Undefined), according to the proportion of pixels that belong to each class. If $70\%$ or more pixels inside the segment belong to Forest or Non-Forest, the volunteer needs to assign this majority class. Otherwise, the volunteer needs to assign as an Undefined class. The segments are presented with different color-compositions, and a tutorial is available to volunteers.

After $15$ or more responses received for each segment from distinct volunteers, the workflow is considered finished, and the contributions are downloaded from the Zooniverse.org platform. These contributions are sent for analysis at the Organization and Selection Module. 

\subsection{Organization and Selection  Module}
\label{metricas}

At the Organization and Selection Module, the class of each segment is defined by the majority vote of the volunteers' answers, and analyses are performed such as the tasks' difficulty level (Section~\ref{sec:entropy}), consensus convergence (Section~\ref{sec:convergence}), and the volunteers' hit rates and score (Section~\ref{sec:score}). 

The Organization and Selection Module is also responsible for the comparison between volunteers' classifications with a given groundtruth when available. For now, PRODES data were used to build different groundtruth that will be explained in Section~\ref{sec:groundtruth}. For the groundtruth, it was needed the definition of a metric called Homogeneity Ratio ($HoR$), presented in Section~\ref{sec:hor}.

With the  classification of the segments, it is possible to create the training set of the Machine Learning Module. Only the segments with classes Forest and Non-Forest are considered, discarding the Undefined segments.
%, which will receive further treatment in the future.

\subsubsection{Homogeneity Ratio ($HoR$)}
\label{sec:hor}
The ForestEyes' volunteers choose between three classes: Forest, Non-Forest, and Undefined. However, a pixel-based groundtruth is built as a binary classification (Forest and Non-Forest). This Non-Forest class encompasses everything that is not Forest, such as deforestation, hydrography, clouds, and others. 

To evaluate the quality of the segments, a Homogeneity Ratio ($HoR$) is defined as the percentage of pixels present inside each segment of the majority class in a binary classification task (Forest and Non-Forest). According to equation~\eqref{HoR}, NFP represents the number of Forest pixels, NNP the Non-Forest pixels' number, and NP the total pixels for a segment ($NP=NFP+NNP$).

\begin{equation}
HoR= \frac{\max(NFP,NNP)}{NP}\label{HoR}
\end{equation}

%To capture the $HoR$ distribution of the segments over the image the following $HoR$ ranges were created and accounting: $0.5 \leq HoR < 0.6$, $0.6 \leq HoR < 0.7$, $0.7 \leq HoR < 0.8$, $0.8 \leq HoR < 0.9$, and  $HoR = 1.0$. Therefore, if a segment has $0.5 \leq HoR < 0.6$ to Forest class means that it has $0.5 \geq HoR$ to Non-Forest class.  

The $HoR$ metric is partitioned in five different ranges to better represent their distribution: $0.5 \leq HoR < 0.6$, $0.6 \leq HoR < 0.7$, $0.7 \leq HoR < 0.8$, $0.8 \leq HoR < 0.9$, and  $HoR = 1.0$. Therefore, if a segment has $0.5 \leq HoR < 0.6$ to Forest class means that it has $0.5 \geq HoR$ to Non-Forest class.

\subsubsection{Groundtruth}
\label{sec:groundtruth}

\begin{figure}[!ht]
\centering
\includegraphics[width=0.7\textwidth,keepaspectratio=true]{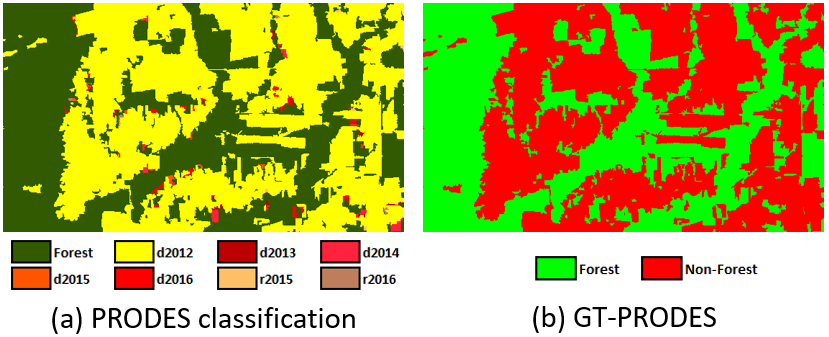}  
\caption{Example of PRODES binarization where the resulting binary image becomes the GT-PRODES. The image presents an area in the State of Rondônia of approximately $64,000$ hectares. } 
\label{fig:images}
\end{figure}

To compare the classification of the volunteers with PRODES data, two strategies were used to define the groundtruth. The first one is pixel-based. Given an image from PRODES mosaic (Figure~\ref{fig:images}(a)), all PRODES' classes different of Forest (deforestation, residue, hydrography, clouds, non-forest vegetation, and others) are defined as Non-Forest, resulting in a binary image (Figure~\ref{fig:images}(b)) that will be referred as GT-PRODES.

The second strategy of groundtruth is segment-based, where two new GTs are created: GT with Undefined (GT-U) and GT with the majority (GT-M). These two GTs are created by considering $HoR$ for each segment. 

\begin{figure}[!ht]
\centering
\includegraphics[width=0.7\textwidth,keepaspectratio=true]{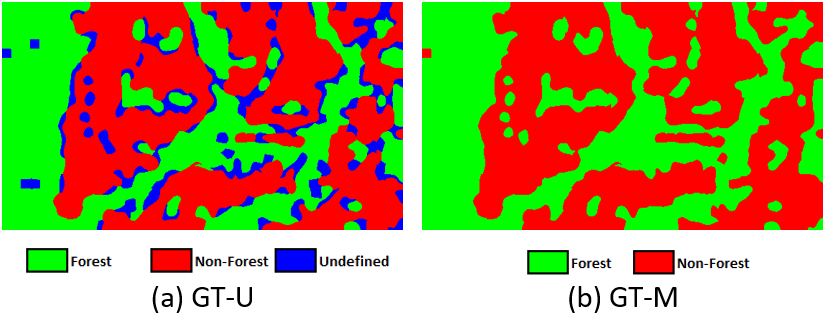} %& 
\caption{Examples of GT-U and GT-M for the same region.}% depicted in Figure~\ref{fig:images}.} 
\label{fig:groundtruth2}
\end{figure}

With GT-U, the segments with $HoR < 0.7$ are classified as Undefined, otherwise ($HoR \geq 0.7$) receive the respective Forest or Non-Forest label. As for GT-M the segment is classified considering its majority class~\cite{foresteyes2019}. Figure~\ref{fig:groundtruth2} presents GT-U and GT-M for the same region depicted in Figure~\ref{fig:images}.

The volunteer classification accuracy is defined here by equation~\eqref{acc}. Let  $NSC$ be the number of samples correctly classified and $TNS$ the total number of samples. For GT-PRODES, the samples are just the pixels, and for GT-U and GT-M, the samples are the segments.

\begin{equation}
    Accuracy = 100 \times \frac{NSC}{TNS} 
    \label{acc}
\end{equation}

\subsubsection{Task's Difficulty Level}
\label{sec:entropy}

According to~\citet{ARCANJO}, the task's difficulty level can be calculated using the volunteers' answers and Shannon's entropy ($HE$)~\cite{SHANNON1}, which is given by equation~\eqref{entropy}. The higher the entropy value, the greater the difficulty of the task.

\begin{equation}
HE= - \sum_{i=1}^{N}p_i \times \log_2 p_i\label{entropy}
\end{equation}

Where $p_i$ is the probability of the class $i$ be chosen, calculated by the ratio between the number of votes given to class $i$ and the total of votes for the task, and $n$ is the number of possible classes in the task.

The tasks' entropy values are normalized to the range $(0,1)$ and empirically separated into three difficulty levels: Easy ($HE$ $ \leq$  $0.33 $); Medium ($0.33 <  HE  \leq 0.66$); and Hard ($HE  > 0.66$).

The average time to the task accomplishment by the volunteers is also taken to be used to measure the difficulty level. This metric could be correlated to entropy, accuracy, or $HoR$, showing different task hardness applied to different volunteer groups and CS campaigns.

\subsubsection{Consensus Convergence}
\label{sec:convergence}
The analysis of the consensus convergence is used to evaluate the number of answers needed in each task to define its final classification. This analysis is performed by checking if the classification given with the predefined number of answers remains the same with fewer answers. For the ForestEyes Project, the first 15 answers define the final classification of each task. Therefore the consensus convergence analysis was considering the achieved classifications for $5$, $7$, $9$, $11$, and $13$ answers.

\subsubsection{Volunteers' Hit Rates and Score}
\label{sec:score}
As in~\citet{ARCANJO}, it is assumed that the answer given by volunteers' consensus is the correct one. The number of hits of each volunteer is incremented by $1$ when the volunteer agrees with the canonical answer and remains unchanged if it doesn't agree. The volunteer's hit rate (HR) and score (VS) are calculated through equations~\eqref{hr} and~\eqref{score}, where hits mean the number of answers the volunteer agreed with the consensus and total\_answers is the number of answers that the volunteer gave.

\begin{equation}
HR=  100 \times \frac{hits}{total\_answers}  \label{hr}
\end{equation}
\begin{equation}
VS= (0.3 \times total\_answers) + (0.7 \times hits) \label{score}
\end{equation}

Besides evaluating the volunteers' answers in comparison with the consensus,~\eqref{hr} can also be used to evaluate the volunteers' answers with the GT bases GT-U and GT-M.

\subsection{Machine Learning Module}

In the Machine Learning Module, a classification algorithm is trained with the set built from the Organization and Selection Module. Then, this algorithm is applied to a test set that consists of segments that were not sent to the Citizen Science Module.

\subsection{Post-processing Module}

The Post-processing Module will evaluate the obtained results from the Machine Learning Module, creating deforestation alerts, for example. Also, it is planned a feedback cycle where test segments will be sent to the Citizen Science Module to improve the training set and consequently improve the classification algorithm.   

%\subsubsection{Metrics and Analysis Tools}
%\label{metricas}

%In this section the metrics and analysis tools applied to ForestEyes Project are explained, emphasizing their purposes. The metrics cover two aspects: image quality and CS evaluation.

\section{Citizen Science's campaigns}
\label{results}

In this section are shown the methodologies and some results of the CS's campaigns. Each subsection describes one single campaign, also known as a workflow in the Zooniverse.org platform, detailing the image used, segmentation algorithms, and user interface.  

\subsection{Beta-Review}

To launch a CS project as a Zooniverse's project, a Beta-Review is needed. The interface is reviewed by volunteers who evaluate the project and provide improvement suggestions. Based on this evaluation, the platform's team defines if the project is ready for an official launch~\cite{foresteyes2019}. 

The ForestEyes' Beta-Review consisted of a workflow with the same tasks as the ForestWatchers' Correct Classification plus six more tasks from Aw\'{a}-Guaj\'{a} as it is needed a minimum of $100$ tasks for a project to be sent for Beta-Review. The volunteers needed to analyze tiles highlighted by red squares with size $3\times3$ pixels from remote sensing images obtained by MODIS sensor, which has a spatial resolution of $250m$. Figure S1 shows the user interface for this Beta-Review.

Contrary to the ForestWatchers Project, which showed the original remote sensing image and the image classified by an automatic procedure, this Beta-Review only showed images from the MODIS sensor. Furthermore, a new answer was created (Undefined class), aiming to identify tasks where there is too much mixture of Forest and Non-Forest pixels (border regions), usually resulting in a low $HoR$.

The Beta-Review received $2,275$ answers being $1,604$ answers from $77$ registered volunteers and $671$ answers from $125$ anonymous volunteers. For Aw\'{a}-Guaj\'{a} $2014$, $31$ tasks were classified as Forest, and just one as Non-Forest whereas for Rond\^{o}nia $2011$, $65$ tasks were classified as Non-Forest, $4$ as Forest and in $3$ tasks there were ties. Table~\ref{tab:entropyFW} presents the frequency of tasks over difficulty levels based on entropy ranges. 

\begin{table}[!ht]
\caption{The frequency of the tasks with difficulty level for the Beta-Review workflow in the ForestEyes Project.}
\centering
\resizebox{0.8\textwidth}{!}{
\begin{tabular}{@{}cccc@{}}
\toprule
% \textbf{Entropy (E) }    & $E  \leq  0.33 $ & $0.33 < E \leq 0.66$ & $E > 0.66$ \\ 
  \textbf{Difficulty Level} & \textbf{Easy (E)}              & \textbf{Medium (M)}                             & \textbf{Hard (H)}                \\ 
  \footnotesize{\textbf{Entropy (\textit{HE})}}   & \footnotesize{$HE \leq 0.33$}           & \footnotesize{$0.33 < HE \leq 0.66$} & \footnotesize{$HE > 0.66$}\\         \midrule
 Awá-Guajá 2014 & 18 (56.25\%)      & 12 (37.5\%)                        & 2 (6.25\%)          \\
 Rondônia 2011  & 4 (5.56\%)        & 26 (36.11\%)                       & 42 (58.33\%)        \\ \bottomrule 
\end{tabular}
}
\label{tab:entropyFW}
\end{table}

Notice that the volunteers had high consensus with Aw\'{a}-Guaj\'{a} tasks, since it was necessary only to compute $5$ answers to achieve the same results as computing the first $15$, and $93.75\%$ of the tasks were considered easy or medium difficulties.
As for Rond\^{o}nia, the majority of tasks ($58.33$\%) were considered hard, demanding $9$ or more answers to achieve a consensus convergence higher than $90$\%.    

Although the Rond\^{o}nia tasks could be considered a challenging task for the volunteers, their classification results were $88.9$\% when compared with PRODES classification (GT-PRODES). As feedback from the volunteers, two problems have been reported: (1) image resolutions, which can be observed in Table~\ref{tab:entropyFW} with a high number of hard tasks; and  (2) how tasks were presented to volunteers. So new workflows were created taking into account their feedback and will be explained in the following.

\subsection{First Official Workflows} 
\label{sec:firstw}

In the first official workflow for the ForestEyes Project at the Zooniverse.org platform, the remote sensing images from the Landsat-8 satellite were used, which allow a better resolution than MODIS sensors. This image covers an area in Rond\^{o}nia in the year of $2016$, and was acquired at EarthExplorer containing $7$ out of $11$ bands. However, this image was resampled from resolution $30m$ to $60m$, matching with the classified image by PRODES~\cite{foresteyes2019}. The size of the generated image is $558 \times 318$ pixels corresponding to an area of approximately $64,000$ hectares.   

Besides using an image with better resolution, a new interface to display the tasks was presented as required by volunteers. So, instead of using fixed square tiles, it was decided to use segments generated by the SLIC algorithm. Therefore, the steps of the Pre-processing Module were developed as explained in Section~\ref{sec:ForestEyes}.

\begin{figure}[!ht]
\centering
\begin{tabular}{cc}
\includegraphics[scale=0.38,keepaspectratio=true]{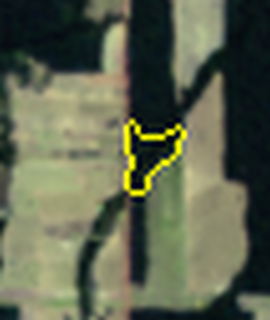} & 
\includegraphics[scale=0.38,keepaspectratio=true]{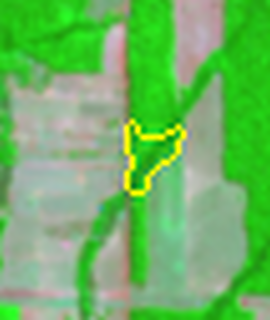} \\
(a) RGB. & (b) $753$ false-color composition. \\
\end{tabular}
\caption{Examples of color compositions available to volunteers.}
\label{fig:compositions}
\end{figure}

For the user interface, presented in Figure S2,  two different color composition images were available as in Figure~\ref{fig:compositions}. In (a) is the original Landsat-8's RGB, and in (b) is a false-color image composed of the bands $7$ (shortwave infrared $2$), $5$ (near-infrared), and $3$ (green) from Landsat-8. This new composition will be referred to as $753$ false-color composition. 

With this new workflow, the ForestEyes Project was officially launched in April $18$, $2019$, where just one week later, all $1,022$ tasks have been completed, receiving $19,803$ answers where $17,239$ answers were from $151$ registered volunteers and the remaining from $76$ anonymous volunteers.

The volunteers classified $1,022$ tasks being $467$ Forest, $467$ Non-Forest, $73$ Undefined, and $15$ tasks tied that were randomly chosen, resulting in a GT-PRODES accuracy of $84.15$\%, GT-U and GT-M accuracies were of $84.25$\% and $86.50$\%, respectively. 

The second workflow was created with an image of the same area acquired in another year ($2013$), where its user interface can be seen in Figure S3. This workflow was also completed in almost a week, reaching $15,610$ answers for $1,028$ tasks, with $137$ registered volunteers giving $14,578$ answers and $41$ anonymous providing $852$ contributions.

These $1,028$ tasks were classified as $478$ being Forest, $470$ Non-Forest, $74$ Undefined, and ties happened in $6$ tasks, resulting in a GT-PRODES accuracy of $83.76$\%, GT-U and GT-M accuracies of $84.34$\% and $88.33$\%, respectively. 

Performed analyses over the results indicated that registered volunteers contributed more than the anonymous. Also, they achieved better accuracies either considering the volunteers' consensus and the groundtruth hit rates (GT HRs), providing more homogeneous answers than the anonymous volunteers. Furthermore, it is possible to observe that the consensus and GT HRs increase as volunteers answered more tasks, stabilizing with a reasonable value~\cite{foresteyes2019}.

Different from Rond\^{o}nia $2011$, where the majority of tasks were hard, in these new workflows with a better image resolution, around $65$\% of the tasks were considered easy, with just $13.8$\% being hard, corresponding to $283$ tasks. Also, the consensus convergence for both workflows ($2016$ and $2013$) achieved more than $90$\% with only $5$ answers.

The comparison of the volunteers' classifications for $2016$ and $2013$ could show the areas where occurred deforestation between those years. Therefore the difference between them was taken and compared with the difference between GT-PRODES $2016$ and GT-PRODES $2013$.

The difference between the two GT-PRODES showed that $2,184$ pixels correspond to deforestation between the years of $2014$ and $2016$. However, from the difference between the two workflows, only $570$ were correctly classified as Non-Forest by the volunteers, $302$ were labeled as Undefined, in $176$ tasks happened ties, and $115$ were wrongfully classified as Forest. 
These results may have happened due to errors in the image segmentation process, failures in the tasks display, variability in the spectral satellite image characteristics, or volunteers' error. 

Intending to get results even closer to GT-PRODES, especially for the detection of areas with recent deforestation, k-means clustering was performed on the tasks considered hard (high entropy), Undefined, and the ties, resulting in a new workflow that is explained next.

\subsection{Refining Segmentation Process through Clustering}
\label{ressegmentacao}

From the results obtained by \citet{foresteyes2019}, and depicted in Section~\ref{sec:firstw}, the Undefined tasks, the tasks with ties, and the hard segments ($HE > 0.66$), received a new segmentation process aiming to refine the polygons/regions. This segmentation turns each original task into two or more samples, which were organized in a new workflow at the Zooniverse.org platform.

\begin{figure}
    \centering
    \includegraphics[width=0.8\textwidth]{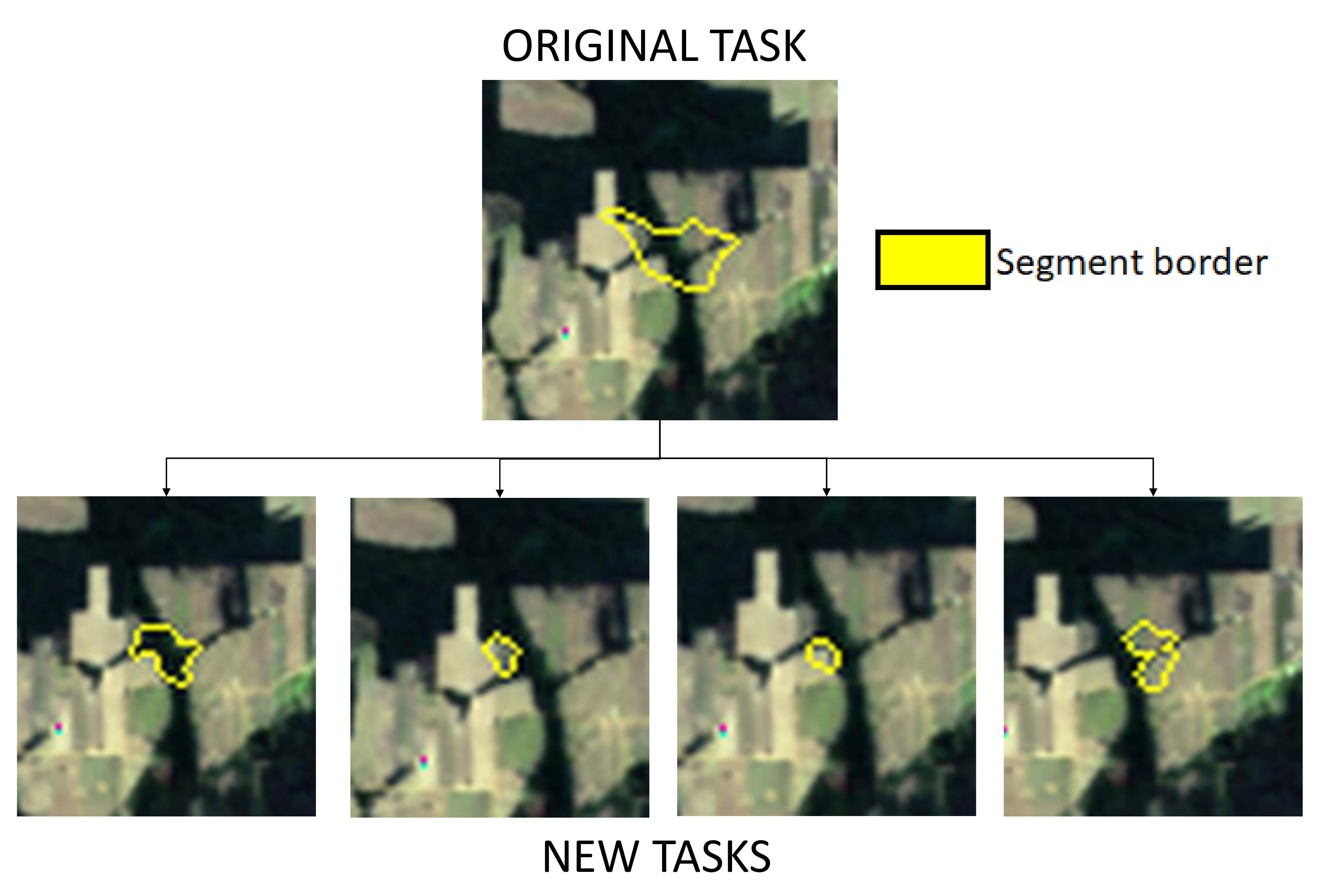}
    \caption{Examples of new tasks for the new workflow.
    The original task was divided into four new segments that were sent to volunteers.}
    \label{fig:newsegments}
\end{figure}

The k-means~\cite{lloyd1982least} algorithm was used to refine the depicted segments, using the PCA in the pre-processed image, and the elbow method~\cite{kodinariya2013review} to define the optimal number of clusters. An example of the new segmentation process can be viewed in Figure~\ref{fig:newsegments}. 

Since the k-means algorithm does not take into account the spatial pixel distribution, distant pixels located outside a new segment are also considered an integrated part of the same cluster.
Therefore, the number of created news segments could be greater than $k$. Some of them could also be too small to visualize.

To make the interface and image analysis more friendly, the segments with less than $9$ pixels were discarded, reaching $1,285$ new tasks sent to volunteers where $637$ are related to $2013$ and $648$ from $2016$. The segments had an average size of $35$ pixels, representing a minor area than the previous workflows, which had $173$ pixels on average.

The user interface, presented in Figure S4, showed two zoomed images side-by-side representing the available color compositions, $753$ false-color and RGB. Also, due to possible problems with the segments' size, a new answer option was available to volunteers ("Segment too small"), making possible a more detailed quality evaluation of the new tasks.

From the $1,285$ tasks and considering the first $15$ answers, $598$ received Forest label, $621$ received Non-Forest, $29$ were classified as Undefined, $7$ was considered too small, and draws happened in $30$ tasks. The workflow received $19,496$ answers, being $18,789$ from $142$ registered volunteers, and $707$ from $40$ anonymous. For better comprehension, the results were divided into the ones regarding $2016$ (defined here as refined$2016$) and the ones from $2013$ (defined here as refined$2013$). 

For refined$2013$, $312$ segments were labeled as Forest, $283$ as Non-Forest, $20$ as Undefined, $4$ as Small, and in $18$ happened ties, resulting in accuracy of $67.50$\% for GT-PRODES, $59.34$\% for GT-U, and $67.19$\% for GT-M. Whereas for refined$2016$, $286$ tasks were classified as Forest, $338$ as Non-Forest, $9$ as Undefined, $3$ as Small, and in $12$ happened ties. For the GT-PRODES accuracy was $71.91$\%, GT-U reaches $62.65$\% and GT-M $72.68$\%.

Although the results didn't achieve satisfying values, being very bellow compared to the other workflows, when integrating these new classifications to the first official workflows described in Section~\ref{sec:firstw}, the previous results from the first workflows regarding GT-PRODES are increased by more than $4$\%. It also increased the monitoring of recent deforestation by approximately $85.9$\%. However, improvements were still required since, even with these optimizations, just $48$\% of GT-PRODES pixels with recent deforestation were found. 

\subsection{New Image Resolution and Segmentation Algorithm}
\label{segmentacao_comparacao}

By $2017$, the PRODES' mosaics started to have a resolution of $30m$, enabling the use of the Landsat-8 images without needing resampling. 
The use of images with better resolution could improve the user visualization and consequently the volunteers' classification.

A Landsat-8 image for a similar region defined for the previous workflows was collected, as close as possible. This image has $1058 \times 625$ pixels, corresponding to an area of approximately $59,500$ hectares, and is the same used to build PRODES mosaic of $2017$.

To improve the segmentation, another algorithm (IFT-SLIC) was tested. This algorithm is an extension of SLIC in an IFT framework~\cite{alexandre2015ift}. Both SLIC and IFT-SLIC were applied to the new remote sensing image, and two new workflows were created at the Zooniverse.org platform: SLIC $2017$ and IFT-SLIC $2017$. Both had the same user interface that can be seen in Figure S5.

By analyzing the $HoR$'s distribution ranges of each algorithm, SLIC presented higher percentage of segments with $HoR = 1.0$ ($43.6$\%) and higher percentage of segments with $HoR \geq 0.7$ ($90.2$\%) compared to IFT-SLIC that had $39.1$\% and $87.1$\%,  respectively.

For SLIC $2017$, $14,339$ answers were received from $64$ registered volunteers and $359$ answers from $19$ anonymous volunteers, resulting in $14,698$ answers for the $973$ tasks, where $417$ were classified as Forest, $450$ as Non-Forest, $104$ as Undefined, and in $2$ tasks happened ties. The accuracy were $83.63$\% for GT-PRODES, $87.87$\% for GT-U, and $87.05$\% for GT-M. With IFT-SLIC $2017$, $15,360$ answers were received, where $14,977$ from $79$ registered volunteers, and $383$ from $20$ anonymous users. The $1,008$ tasks were classified with $448$ as being Forest, $443$ as Non-Forest, and $117$ Undefined, resulting in a GT-PRODES accuracy of $81.39$\%, GT-U of $86.31$\%, and GT-M of $85.62$\%.  

SLIC $2017$ presented better results regarding all GTs compared with IFT-SLIC $2017$. This could be explained by SLIC's higher percentage of homogeneous segments ($HoR \geq 0.7$ and $HoR = 1.0$), which may have helped the volunteers in the classification task.

\subsection{Improving the Detection of Areas with Recent Deforestation}
\label{maskslic}

As the project proposes to monitor rainforests' deforestation, detecting areas with recent deforestation is the most important challenge to overcome. As seen in the refined workflow (Section~\ref{ressegmentacao}), only $48$\% of recent deforestation was found, requiring an enhancement process. 

Firstly, to exclusively monitor areas with recent deforestation, it is needed to segment images only in non-consolidated deforestation areas detected by PRODES or deforested areas undetected by previous campaigns.

In this new workflow, a segmentation algorithm called\\ MaskSLIC~\cite{irving2016maskslic}, which allows to segment regions outside a mask over the image, was adopted. 
A mask from the PRODES $2017$ image was created, taking into account only a period of deforestation before August/$2016$. These pixels with recent deforestation (encompassing the deforestation between August/$2016$ and July/$2017$) were the targets to be detected in this new campaign.

The number of segments was defined as the number of pixels outside the mask divided by $70$ pixels, which approximately corresponds to the minimum size area of deforestation monitored by PRODES ($6.25$ hectares)~\cite{gomes2014amazon}. So, the final number of segments was $4,466$. 

As sending all these segments to the volunteers would be impractical, an analysis using the groundtruth and $HoR$ values was performed to build this new workflow. 
Hence, all the Non-Forest and Forest segments with $HoR < 1.0$, and some Forest segments with $HoR = 1.0$ were selected, resulting in $92$ tasks.

\begin{figure*}[!ht]
\centering
%\begin{tabular}{ccc}
%\includegraphics[scale=0.25,keepaspectratio=true]{2017_recent_753task_4134.png} &
%\includegraphics[scale=0.25,keepaspectratio=true]{2017_recent_RGBtask_4134.png} & 
%\includegraphics[scale=0.25,keepaspectratio=true]{2017_recent_NDVItask_41342.png} \\
%\includegraphics[scale=0.5,keepaspectratio=true]{legend_border.png}\\
%(a) $753$ false-color composition.  & (b) RGB composition.  & (c) NDVI image.  \\
%\end{tabular}
\includegraphics[width=0.8\textwidth]{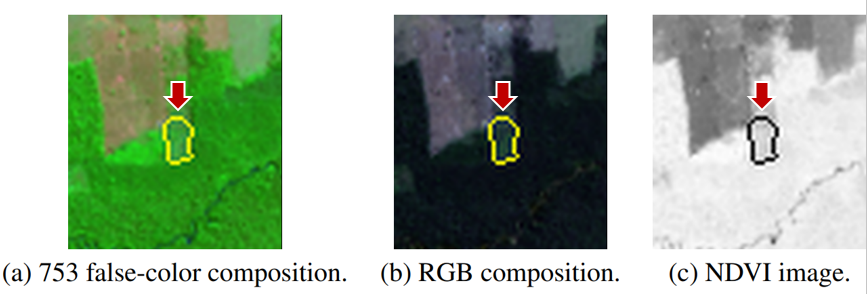}
\caption{Example of task for the new ForestEyes's workflow with MaskSLIC. The images show a recent deforested area.} 
\label{fig:maskslicexample}
\end{figure*}

Remaining vegetation are commonly found in the Amazon deforestation process, which hinders the recent deforestation detection~\cite{rs12060910} not only for automatic algorithms but also for people.
To help the volunteers' classification task, an NDVI~\cite{guide2017landsat} image was also displayed in the interface.
This vegetation index is calculated as a ratio between bands red and NIR, given by Equation~\eqref{ndvi}. It results in a grayscale image, where the darkest pixels can be water, clouds, rock, or sand; moderate pixels can be grassland or shrub, and the brightest pixels (closest to white) represent dense green leaves. 
Figure~\ref{fig:maskslicexample} shows an example of images displayed to volunteers in this new ForestEyes' workflow with the MaskSLIC algorithm, representing an area with recent deforestation, and Figure S6 presents the user interface of this workflow.  

\begin{equation}
NDVI= \frac{NIR - Red}{NIR + Red}  \label{ndvi}
\end{equation}

This new workflow received $1,481$ answers from $28$ volunteers where $1,476$ answers from $26$ registered volunteers and $5$ answers from $2$ anonymous. In a total of $92$ tasks, $53$ were classified as Forest, $23$ as Non-Forest, and $16$ as Undefined. The resulting accuracies were $75.65$\% for GT-PRODES, $85.87$\% for GT-U, and $80.43$\% for GT-M.

Considering just recent deforested areas, the scheme presented in this subsection shows an improvement in the detection of areas with recent deforestation achieving $65.95$\% for GT-PRODES accuracy wherein the previous workflows reached only $48$\%. However, the images taken for both workflows are not the same, differing by years, for almost the same coverage area.

\section{Comparison Between Workflows and Discussions}
\label{sec:comparison}
This section shows different comparative analyses between workflows.
Firstly, the general information is presented regarding the number of tasks, volunteers, and answers for each workflow. 
Next, accounting for the workflows tasks' contributions, the consensus convergence, and CS's accuracy for the different GTs are evaluated and analyzed.
The tasks' difficulty levels by entropy and answers' average time are also calculated and correlated. Finally, the volunteers' ranking is built according to the different scores computed from all workflows.   

\subsection{General Information}

Table~\ref{tab:generalinfo} shows the general information for the ForestEyes' official workflows, where the registered volunteers had significant participation, even reaching $295$ times greater than the anonymous answers for the MaskSLIC workflow. 

\begin{table*}[!ht]
\caption{Summary of the ForestEyes' workflows.}

\centering
\begin{tabular}{@{}cccc|ccc@{}}
\toprule
\multirow{2}{*}{\textbf{Workflow}} & \multirow{2}{*}{\textbf{2013}} & \multirow{2}{*}{\textbf{2016}} & \multirow{2}{*}{\begin{tabular}[c]{@{}c@{}}\textbf{refined}\\ \textbf{2013 \& 2016}\end{tabular}} & \multicolumn{3}{c}{\textbf{2017}} \\
&                       &                       &                                                                                 & \textbf{SLIC}    & \textbf{IFT-SLIC}    & \textbf{MaskSLIC}  \\ \midrule
Number of Tasks                                                                    & 1,028                  & 1,022                  & 1,285                                                                            & 973     & 1,008   & 92    \\ \midrule \midrule
Registered volunteers                                                    & 137                   & 151                   & 102                                                                             & 64      & 79     & 26    \\
Anonymous volunteers                                                     & 41                    & 76                    & 40                                                                              & 19      & 20     & 2     \\ 
\textbf{Total of volunteers}   & 178&	227	&142&	83&	99&	28    \\ 

\midrule \midrule

\begin{tabular}[c]{@{}c@{}}Registered Answers\\ \end{tabular} & 14,758                 & 17,239                 & 18,789                                                                           & 14,339   & 14,977  & 1,476  \\
\begin{tabular}[c]{@{}c@{}}Anonymous Answers\\ \end{tabular}  & 852                   & 2,564                  & 707                                                                             & 359     & 383    & 5  \\ %\midrule  
\textbf{Total of Answers}                                                        & 15,610                 & 19,803                 & 19,496                                                                           & 14,698   & 15,360  & 1,481  \\ \bottomrule
\end{tabular}
\label{tab:generalinfo}
\end{table*}

For the next results, the refined workflow $2013\& 2016$ was divided into the tasks from $2013$ (refined$2013$) and the tasks from $2016$ (refined$2016$). 

\subsection{Consensus Convergence}

The consensus convergence by the number of redundant contributions, defined in Section~\ref{sec:convergence}, was taken for the ForestEyes' official workflows, resulting in Table~\ref{table:convergence}. For all workflows, except refined$2013$, the convergence for just $5$ answers is higher than $90$\%. For SLIC and IFT-SLIC, almost $97$\% of the tasks have the same classification with $5$ and $15$ answers. Furthermore, only MaskSLIC presents a decrease when the considered number of answers reaches $13$, but the convergence's percentage remains high, with almost $95$\%. 
 
For future workflows is planned to use fewer answers than the predefined $15$. The first attempt will be with $11$ answers. 
 
\begin{table}[!ht]
\caption{Consensus Convergence for ForestEyes' workflows.}
\centering
\resizebox{0.8\textwidth}{!}{
\begin{tabular}{@{}ccccccc@{}}
\toprule
 \multirow{2}{*}{\textbf{Workflows}}& \multicolumn{5}{c}{\textbf{Number of Answers}} \\ \cline{2-6}

& \textbf{5} & \textbf{7}       & \textbf{9}       & \textbf{11}      & \textbf{13}      \\ \midrule
 2013              & 94.75\% & 95.72\% & 96.00\% & 97.47\% & 98.54\% \\
                            2016              & 90.50\% & 93.44\% & 95.30\% & 96.80\% & 98.14\% \\
                            refined2013       & 88.23\% & 90.11\% & 90.11\% & 93.41\% & 96.08\% \\
                            refined2016       & 91.36\% & 94.44\% & 95.68\% & 96.60\% & 97.84\% \\ \hline
                            SLIC          & 96.81\% & 97.74\% & 98.15\% & 99.08\% & 99.08\% \\
                            IFT-SLIC      & 97.32\% & 98.21\% & 98.51\% & 99.21\% & 99.60\%   \\
                            MaskSLIC      & 91.30\% & 93.48\% & 95.65\% & 96.74\% & 94.56\% \\ \bottomrule 
\end{tabular}
}
\label{table:convergence}
\end{table}

\subsection{Citizen Science's Accuracy}

With the $3$ GTs (pixel-based GT-PRODES, segment-based GT-U and GT-M) defined in Section~\ref{sec:groundtruth}, the classifications performed by ForestEyes' volunteers achieved the results presented in Table~\ref{table:accuracy}.

%Table~\ref{table:accuracy} shows the overall accuracy for each workflow and GT.

\begin{table}[!ht]
\centering
\caption{Overall accuracies for each workflow and GT in the ForestEyes Project.}
\begin{tabular}{@{}cccccc@{}}
\toprule
\textbf{Resolution} &   \textbf{Workflows} & 
\textbf{GT-PRODES} & \textbf{GT-U}  & \textbf{GT-M} \\ 
&& \scriptsize{(Pixel-based)} & \multicolumn{2}{c}{\scriptsize{(Segment-based)}} \\                          
                        \midrule
                        \multirow{4}{*}{$60m$}
                      &  2013          & 83.76\% & 84.34\% & 88.33\%  \\
                     &       2016          & 84.15\% & 84.25\% &86.50\%  \\ %\midrule%\cline{2-5}
                      &      refined2013   & 67.50\% & 59.34\% & 67.19\%  \\
                      &      refined2016   & 71.91\% & 62.65\% & 72.68\% \\ \midrule
                   \multirow{3}{*}{$30m$}   &      SLIC      & 83.63\% & 87.87\% & 87.05\%  \\
                      &      IFT-SLIC  & 81.39\% & 86.31\% & 85.62\%  \\
                \cline{2-5}      &      MaskSLIC  & 75.65\% & 85.87\% & 80.43\%  \\ \bottomrule 
\end{tabular}
\label{table:accuracy}
\end{table}

The workflows with $60m$ resolution images ($2013$, $2016$ and the refined workflow) have GT-M accuracy better than GT-U, which is explained by their poor accuracy for Undefined segments, reaching only $27.94$\% for the $2016$ workflow. 
Maybe the volunteers just looked the simple majority class into the segment.

The Undefined class presented better accuracy for the workflows from the $2017$ remote sensing image ($30m$ resolution), especially to MaskSLIC workflow, where it achieved a reasonable value of $61.54$\%. 
However, compared to the other classes (Forest or Non-Forest), its accuracy is still low. 
The complex segments needing refinement or adjustments usually demand an undefined classification, but this situation could not be perceived and handled.
Investigations to identify Undefined segments before they are sent to volunteers are ongoing, enhancing the image segmentation, which will create better tasks to be sent to volunteers. 

The Forest class had the best accuracy for all workflows and GTs, being above $88$\% for most cases. For GT-PRODES, although MaskSLIC had the worst accuracy compared to the $2013$, $2016$, SLIC, and IFT-SLIC workflows, it achieved $65.95$\% accuracy for Non-Forest class which in this case represent pixels for recent deforestation. 
The areas with recent deforestation detected by previous workflows reached only $26$\% for the difference between the original $2016$ and $2013$. By using refined segments (Section~\ref{ressegmentacao}) it increased to $48$\%. 
So the accuracy was significantly improved applying the last workflow's methodology, although the years and image resolution are different.  
For future workflows is advisable to use MaskSLIC instead of getting the difference between two images segmented by SLIC. However, a better segmentation segment method will continue to be sought.

%The MaskSLIC workflow, as PRODES~\cite{PRODES3}, does not allow the detection of regenerated areas when a former deforested area became forest again since all consolidated deforestation area is covered by a mask. For that a new workflow with a different methodology to detect only regenerated areas needs to be developed. 

%However, the CS's accuracy would not drop when the groundtruth keeps classifying these areas as deforested, in opposite to the volunteer classifications, as occurs in the other workflows.

\subsection{Task's Difficulty Level}

With the definition of entropy ($HE$) and its classification, given in Section~\ref{sec:entropy}, Table~\ref{tab:entropias} shows the difficulty level frequency for each workflow. 
Although IFT-SLIC seems to be easier for volunteers, SLIC was the one that achieved the best results for all three GTs (see in Table~\ref{table:accuracy}). 

The workflows with resolution images of $30m$ show a desirable decrease in the proportion of hard tasks. However, MaskSLIC didn't present the same magnitude for decreasing as the other two workflows. Also, it presented low GT accuracy when compared to the former workflows, showing how challenging can be the detection of recent deforestation, even with NDVI's image.

Another difference between MaskSLIC and the other $2017$ workflows methodology refers to the segments' size. In MaskSLIC, the segments had an average size of $70$ pixels, which could have affected the visualization and the pattern recognition by the volunteers. Remembering a similar issue occurred with the refined segments (Section~\ref{ressegmentacao}).

\begin{table}[!ht]
\centering
\caption{Tasks' difficulty level frequency for ForestEyes' workflows.}
\resizebox{1.00\linewidth}{!}{
\begin{tabular}{@{}cccc@{}}
\toprule
                                  
                           % \textbf{Difficulty Level}& \textbf{Easy}    & \textbf{Medium} &  \textbf{Hard}    \\ 
                           % \footnotesize{\textbf{Entropy (\textit{E})}}   & \footnotesize{$E \leq 0.33$}           & \footnotesize{$0.33 < E \leq 0.66$} & \footnotesize{$E > 0.66$}          \\
                           \textbf{Difficulty Level} & \textbf{Easy (E)}              & \textbf{Medium (M)}                             & \textbf{Hard (H)}                \\ 
  \footnotesize{\textbf{Entropy (\textit{HE})}}   & \footnotesize{$HE \leq 0.33$}           & \footnotesize{$0.33 < HE \leq 0.66$} & \footnotesize{$HE > 0.66$}\\         
                            \midrule
 2013           & 681 (66.25\%) & 205 (19.94\%)         & 142 (13.81\%) \\
                            2016           & 648 (63.40\%) & 233 (22.80\%)         & 141 (13.80\%) \\
                            refined2013    & 365 (57.30\%) & 161 (25.27\%)          & 111 (17.43\%)   \\
                            refined2016    & 457 (70.52\%) & 99 (15.28\%)          & 92 (14.20\%)   \\ \hline
                            SLIC       & 665 (68.35\%) & 267 (27.44\%)          &  41 (4.21\%)      \\
                            IFT-SLIC   & 734 (72.82\%) & 241 (23.91\%) & 33 (3.27\%)       \\
                            MaskSLIC   & 59 (64.13\%) & 23 (25.00\%)           &   10 (10.87\%)     \\ \bottomrule 
\end{tabular}
}
\label{tab:entropias}
\end{table}

Table~\ref{tab:entropy_class} shows the difficulty level by each class. It is possible to verify that Undefined tasks were the hardest ones, followed by the Non-Forest tasks, being Forest segments the easiest ones. For workflows with $60m$ resolution, the vast majority of Undefined segments were difficult. However, for workflows with $30m$ resolution, the great majority of Undefined tasks were easy or of medium difficulty, which can explain the improvement of Undefined and GT-U accuracies. 

For MaskSLIC, most Non-Forest segments were hard or of medium difficulty, contrary to the findings with SLIC and IFT-SLIC workflows. This reinforces how challenging can be the detection of recent deforestation. 

\begin{table*}[!ht]
\caption{Tasks’ difficulty level frequency for each class for ForestEyes' workflows. The Forest class is defined as F, Non-Forest as NF, Undefined as U, and Small as S.}
\resizebox{1.15\linewidth}{!}{
\begin{tabular}{@{}cccc@{}}
\toprule
\textbf{Difficulty Level} & \textbf{Easy (E)}              & \textbf{Medium (M)}                             & \textbf{Hard (H)}                \\ 
  \footnotesize{\textbf{Entropy (\textit{HE})}}   & \footnotesize{$HE \leq 0.33$}           & \footnotesize{$0.33 < HE \leq 0.66$} & \footnotesize{$HE > 0.66$}\\         \midrule
2013                    & 84.10\%F/59.36\%NF          & 9.83\%F/30.43\%NF/20.27\%U           & 6.07\%F/10.21\%NF/79.73\%U           \\
2016                    & 81.58\%F/56.74\%NF/2.74\%U  & 10.92\%F/35.12\%NF/24.66\%U          & 7.50\%F/8.14\%NF/72.60\%U            \\
refined2013             & 68.59\%F/53.36\%NF          & 19.23\%F/33.21\%NF/20.00\%U/50.00\%S & 12.28\%F/13.43\%NF/80.00\%U/50.00\%S \\
refined2016             & 81.47\%F/66.27\%NF          & 10.49\%F/19.53\%NF                   & 8.04\%F/14.20\%NF/100\%U/100\%S      \\ \hline
SLIC                    & 82.49\%F/65.11\%NF/26.92\%U & 16.07\%F/31.33\%NF/56.73\%U          & 1.44\%F/3.56\%NF/16.35\%U            \\
IFT-SLIC                & 82.14\%F/74.94\%NF/29.06\%U & 15.85\%F/23.25\%NF/57.26\%U          & 2.01\%F/1.81\%NF/13.68\%U            \\
MaskSLIC                & 81.13\%F/43.48\%NF/37.50\%U & 13.21\%F/39.13\%NF/43.75\%U          & 5.66\%F/17.39\%NF/18.75\%U           \\ \bottomrule
\end{tabular}
}
\label{tab:entropy_class}
\end{table*}

\subsection{Tasks' Time}

The Zooniverse.org platform registers logs for each answer, which define the date and time for the initial and final steps.
Therefore, the time spent on each answer was evaluated and filtered to eliminate some outliers. The average times were calculated regarding all workflows' answers.
Measurements were taken for different categories, including the various tasks classes (Forest as F, Non-Forest as NF, Undefined as U and Small as S), the volunteers' type (registered as R and anonymous as A), and the difficulty levels (easy as E, medium as M and hard as H). All these results are presented in Table~\ref{tab:times}.

\begin{table*}[!ht]
\caption{Average time (in seconds) regarding different categories. The classes Forest, Non-Forest, Undefined, and Small are defined as F, NF, U, and S. The registered and anonymous volunteers are defined as R and A. The difficulty levels Easy, Medium, and Hard are defined as E, M, and H.}
\centering
\begin{tabular}{@{}ccccc@{}}
\toprule
  \multirow{2}{*}{\textbf{Workflows}}            & \multicolumn{4}{c}{\textbf{Average Time (s)}}                                     \\ 
     & \textbf{General} & \textbf{Classification}           & \textbf{Volunteers}  & \textbf{Difficulty Level}            \\ \midrule
2013          & 5.83    & 5.15F/5.85NF/9.06U       & 5.76R/7.19A & 5.06E/6.80M/8.19H  \\
2016          & 6.96    & 6.08F/6.94NF/10.48U      & 6.56R/9.75A & 5.80E/8.33M/10.07H \\
refined2013   & 3.84    & 3.57F/3.42NF/6.44U/5.79S & 3.75R/6.00A & 3.13E/4.45M/5.29H  \\
refined2016   & 3.49    & 3.34F/3.07NF/6.25U/6.44S & 3.41R/5.69A & 3.10E/3.95M/4.95H  \\ \hline
SLIC      & 3.29    & 2.91F/3.15NF/4.78U       & 3.23R/5.72A & 2.86E/4.12M/4.87H  \\
IFT-SLIC  & 3.54    & 3.34F/3.21NF/4.86U       & 3.51R/4.99A & 3.16E/4.50M/5.11H  \\
MaskSLIC  & 5.31    & 4.84F/5.73NF/6.07U/6.36S & 5.30R/8.33A & 4.30E/6.40M/8.71H   \\ \bottomrule
\end{tabular}
\label{tab:times}
\end{table*}

The answers that received Undefined or Small classification, anonymous volunteers, and hard tasks took longer. Considering the difficulty level is possible to see that the average time increases as the difficulty increases, once tasks with medium difficulty took longer than the easy ones, and the hardest took longer than the medium.

The MaskSLIC workflow had an average time greater than SLIC, which has the same image resolution. This could have happened due to a small segment size in MaskSLIC, joined to user interface characteristics, since it requires analysis of three images instead of two as the other workflows. Also, from Table~\ref{tab:entropias} this last workflow presented a percentage of hard tasks closer to the $2013$ and $2016$ workflows than the other workflows with $30m$ resolution (SLIC and IFT-SLIC). 

The $2016$ workflow may have had a longer average time because the volunteers had to flip the images to see both color compositions while the other workflows displayed the color compositions side-by-side as default. Both $2016$ and $2013$ workflows also didn't have a default zoom for the tasks, different from the subsequent workflows. Therefore, the volunteers needed to zoom in to get closer to the tasks, which could dispense some time. These conditions could explain why $2016$ and $2013$ workflows have almost $6$s/$7$s as average time while the new tasks from refined$2016$ and refined$2013$ spent only $4$s. 
%\textcolor{red}{Aqui poderia referenciar as imagens das interfaces no material supplementar?}

\subsection{Accuracy Regarding Task's Difficulty Level and $HoR$}

\begin{figure*}[!ht]
   \centering
    \resizebox{1.152\textwidth}{!}{\includegraphics[width=\textwidth]{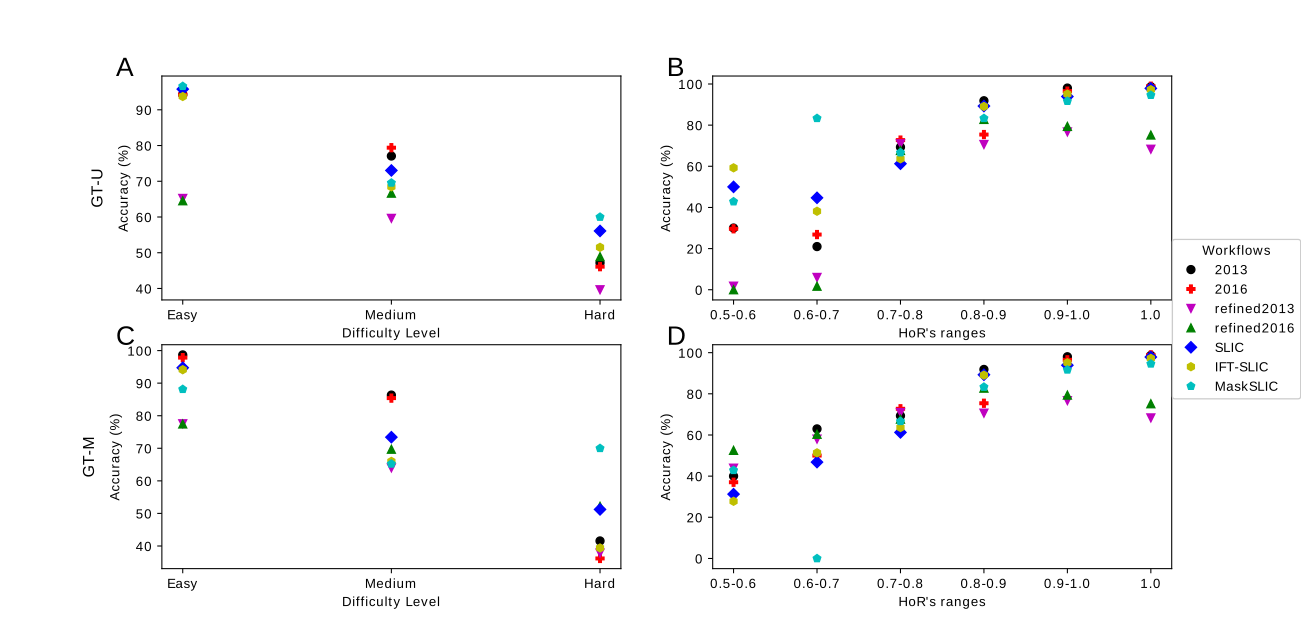}}
    \caption{Accuracies for each workflow considering difficulty level and $HoR$.}
    \label{fig:entropy_hr}
\end{figure*}

Figure~\ref{fig:entropy_hr} shows the accuracy for different GT (GT-U and GT-M) by considering the tasks' difficulty level and $HoR$. It can be seen that for both GTs, the accuracies drop as the difficulty arises for most workflows (Figures~\ref{fig:entropy_hr} A and C). 
As for $HoR$, for GT-M (Figure~\ref{fig:entropy_hr} D), the accuracy increases as the segments are more homogeneous, except for MaskSLIC once it had a reasonable Undefined class accuracy, especially with segments with $HoR$ between $0.6$ and $0.7$.

For GT-U (Figure~\ref{fig:entropy_hr} B), the volunteers had better accuracy for segments with $HoR$ below $0.6$ than segments with $HoR$ between $0.6$ and $0.7$ once as less homogeneous the segment as better the volunteers classify them as Undefined. However, the accuracy is still low, reinforcing the need to early notice Undefined segments, to perform segmentation refinements, and only after that, send them to the volunteers.

This behavior does not occur for MaskSLIC, wherein GT-U accuracy for $HoR$ between $0.6$ and $0.7$ was much higher than the first range ($HoR$ below $0.6$). This could have happened due to these tasks' visualization (i.e. color or distribution of the pixels and segments' shapes), which enabled the volunteers to notice the mixture between Forest and Non-Forest pixels. More investigation is planned to understand this abnormal outcome.

\subsection{Volunteers' Score}

The volunteers' scores, according to the metrics defined in Section~\ref{sec:score}, were evaluated, generating a volunteers' ranking where the first five positions are shown in Table~\ref{pontuacaoRegFE}. The consensus hits at the third column show the number of contributions in agreement to the consensus. The hit rate ($HR$) consensus represents the same value as a relative measurement. The fifth column presents the Volunteer Score ($VS$) value. The last two columns show the HR related to GT-U and GT-M.

\begin{table*}[!ht]
 \centering
 \caption{Volunteers' ranking for ForestEyes Project.}
 \begin{tabular}{ccccccc}
 \toprule
 \textbf{User\_id}&\begin{tabular}[x]{@{}c@{}}\textbf{Number}\\\textbf{answers}\end{tabular}&\begin{tabular}[x]{@{}c@{}}\textbf{Consensus}\\ \textbf{hits}\end{tabular}&\begin{tabular}[x]{@{}c@{}}\textbf{$HR$}\\\textbf{consensus}\end{tabular} &\textbf{$VS$}&\begin{tabular}[x]{@{}c@{}}\textbf{$HR$}\\\textbf{GT-U} \end{tabular} &\begin{tabular}[x]{@{}c@{}}\textbf{$HR$}\\\textbf{GT-M} \end{tabular}\\ \midrule
1805790 & 4,459 & 4,194 & 94.06\% & 4273.5 & 86.61\% & 86.41\%\\
1917133 & 3,255 & 3,065 & 94.16\% & 3122   & 78.59\% & 82.33\%\\
1756349 & 3,266 & 2,950 & 90.32\% & 3044.8 & 71.28\% & 73.48\%\\
1733572 & 3,266 & 2,860 & 87.57\% & 2981.8 & 77.56\% & 78.44\%\\
1061480 & 2,633 & 2,480 & 94.19\% & 2525.9 & 86.14\% & 90.32\%\\
...&...&...&...&...&...&...\\ \bottomrule
 \end{tabular}
 \label{pontuacaoRegFE}
\end{table*}

By dividing the volunteers into registered and anonymous, the average $HR$s for consensus, GT-U, and GT-M were calculated and are presented in Table~\ref{tab:avgvolunteers}. Besides the registered volunteers contributing more than the anonymous, as seen in Table~\ref{tab:generalinfo}, they also achieve better performance.

\begin{table}[!ht]
\centering
\caption{Average HRs for registered and anonymous volunteers.}
\label{meanCHT}
\begin{tabular}{@{}ccc@{}}
\toprule
        \textbf{Average $HR$} & \textbf{Registered} & \textbf{Anonymous} \\ \midrule
Consensus  & 79.22\%    & 75.62\%   \\
GT-U  & 69.21\%    & 63.13\%   \\
GT-M  & 72.24\%    & 69.03\%   \\\bottomrule
\end{tabular}
\label{tab:avgvolunteers}
\end{table}

\section{Conclusion}
\label{conclusao}

This paper described the ForestEyes Project and the steps taken to attest its feasibility regarding the volunteers' contributions which encompassed the following modules: Pre-processing, Citizen Science, and Organization and Selection. The remote sensing images used were from Landsat-8 satellites, and PRODES-based data were used as groundtruth.

The six official CS campaigns covered variations in image resolution ($60m$ and $30m$), segmentation algorithm (SLIC, IFT-SLIC, and MaskSLIC), and user interfaces.
More than $86,000$ contributions were received from $644$ distinct volunteers, being $174$ anonymous and $470$ registered. The registered volunteers contributed much more than the anonymous and achieved higher average HRs for both consensus and GTs.

The upgrade of the image resolution combined with the user interface enhancement seems to improve the volunteers' contributions, reaching higher accuracy regarding the Undefined class and decreasing the percentage of hard tasks. However, more improvements are needed to identify Undefined segments before sending them to volunteers. The recognition of Undefined segments is a very important issue as it enables the refinement of these segments and thus improves the volunteers' accuracy regarding GT-PRODES. 

%\textcolor{red}{falar do por que se deve utilizar MaskSLIC}

%\textcolor{blue}{Seria necessário?}

The good results obtained in this paper show the power of the CS and the Zooniverse.org community. The continuous improvements in the ForestEyes Project could enable the use of CS data as a possible complement to official forest monitoring programs, observing recently deforested areas and contributing to the preservation of the world's biggest cradle of biodiversity.

\section{Future Work}
\label{future}

As future work to turn the ForestEyes Project into an operational system, some challenges need to be overcome, such as:

\begin{itemize}
    \item Pre-Processing Module: new segmentation methods (for example, Deep Learning for semantic segmentation~\cite{long2015fully}), automatic pre-identification of Undefined segments, processing to be performed with new types of Non-Forest (as clouds and rivers).
    \item Citizen Science Module: the selection of segments to be sent to the volunteers, less dependence on PRODES classification, investigations into the volunteers' perceptions or cognition, and visualization enhancements.
    \item Machine Learning Module: the development of this module is ongoing, where the volunteers' classifications will be used with a machine learning paradigm, known as Active Learning (AL)~\cite{TUIA2011}, to build a robust training set. A previous work~\cite{DALLAQUA} showed the feasibility of using AL with PRODES data, where the training set was incremented with pixels.  This new training set will be composed of segments, so a new type of representation is required, wherein Haralick texture features~\cite{HARALICK} and deep learning features can be applied.
    \item Next campaigns: to check the deforestation pattern in other forest areas, new geographical regions, new satellites, and to apply this project to other application domain as the use and occupancy of urban land in hillsides in the fountainhead protection area.
  
\end{itemize}
%on the CS phase as the selection of segments to be sent to the volunteers, automatic pre-identification of Undefined segments, segmentation refinements, less dependence on PRODES, processing to be performed with new types of Non-Forest (as clouds and rivers), investigations into the volunteers' perceptions/cognition, and visualization enhancements.

%It was clear the need for better segmentation, to improve accuracy related to areas with recent deforestation. One investigation that is planned is to use Deep Learning for semantic segmentation~\cite{long2015fully}.
%Future campaigns are intent on checking for the deforestation pattern in other forest areas, like new geographical regions and the use and occupancy of urban land in hillsides in the fountainhead protection area.

%To complete the project, the development of the ForestEyes machine learning module is ongoing, where the volunteers' classifications will be used with a machine learning paradigm known as Active Learning (AL)~\cite{TUIA2011} to build a robust training set.
%The previous work~\cite{DALLAQUA} showed the feasibility of using AL with PRODES data, where the training set was incremented with pixels.  

%As this new training set will be formed by segments of pixels, a new type of representation is required, where Haralick texture features~\cite{HARALICK} will be used.

\section*{Acknowledgment}

The authors would like to thank the research funding agencies CAPES (scholarship), CNPq through the Universal Project (grant \#408919/2016-7). This research is part of the INCT of the Future Internet for Smart Cities funded by CNPq (grant \#465446/2014-0),
CAPES (grant \#88887.136422/2017-00), and FAPESP (grants \#2014/50937-1, \#2015/24485-9, and \#2018/23908-1). Also, the authors would like to thank the ForestWatchers team for the data, INPE for PRODES data, U.S. Geological Survey for the Landsat-8 imagery, the Zooniverse.org team for hosting ForestEyes, and the volunteers who helped in the tasks and performed this study possible. This publication uses data generated via the Zooniverse.org platform, the development of which is funded by generous support, including a Global Impact Award from Google and by a grant from the Alfred P. Sloan Foundation.

\bibliography{mybibfile}

\begin{thebibliography}{72}
\providecommand{\natexlab}[1]{#1}
\providecommand{\url}[1]{\texttt{#1}}
\expandafter\ifx\csname urlstyle\endcsname\relax
  \providecommand{\doi}[1]{doi: #1}\else
  \providecommand{\doi}{doi: \begingroup \urlstyle{rm}\Url}\fi

\bibitem[Achanta et~al.(2012)Achanta, Shaji, Smith, Lucchi, Fua, and
  S{\"u}sstrunk]{SLIC}
Radhakrishna Achanta, Appu Shaji, Kevin Smith, Aurelien Lucchi, Pascal Fua, and
  Sabine S{\"u}sstrunk.
\newblock Slic superpixels compared to state-of-the-art superpixel methods.
\newblock \emph{IEEE transactions on pattern analysis and machine
  intelligence}, 34\penalty0 (11):\penalty0 2274--2282, 2012.

\bibitem[Adam-Bourdarios et~al.(2015)Adam-Bourdarios, Cameron,
  Filip{\v{c}}i{\v{c}}, Lancon, Wu, et~al.]{ATLAS}
C~Adam-Bourdarios, D~Cameron, A~Filip{\v{c}}i{\v{c}}, E~Lancon, Wenjing Wu,
  et~al.
\newblock Atlas@ home: harnessing volunteer computing for hep.
\newblock In \emph{Journal of Physics: Conference Series}, volume 664, page
  022009. IOP Publishing, 2015.

\bibitem[Afelt et~al.(2018)Afelt, Frutos, and Devaux]{afelt2018bats}
Aneta Afelt, Roger Frutos, and Christian Devaux.
\newblock Bats, coronaviruses, and deforestation: {T}oward the emergence of
  novel infectious diseases?
\newblock \emph{Frontiers in microbiology}, 9:\penalty0 702, 2018.

\bibitem[Alexandre et~al.(2015)Alexandre, Chowdhury, Falcao, and
  Miranda]{alexandre2015ift}
Eduardo~Barreto Alexandre, Ananda~Shankar Chowdhury, Alexandre~Xavier Falcao,
  and Paulo A~Vechiatto Miranda.
\newblock Ift-slic: A general framework for superpixel generation based on
  simple linear iterative clustering and image foresting transform.
\newblock In \emph{2015 28th SIBGRAPI Conference on Graphics, Patterns and
  Images}, pages 337--344. IEEE, 2015.

\bibitem[Almeida et~al.(2016)Almeida, Coutinho, Esquerdo, Adami, Venturieri,
  Diniz, Dessay, Durieux, and Gomes]{almeida2016high}
Cl{\'a}udio Aparecido~de Almeida, Alexandre~Camargo Coutinho, J{\'u}lio
  C{\'e}sar Dalla~Mora Esquerdo, Marcos Adami, Adriano Venturieri,
  Cesar~Guerreiro Diniz, Nadine Dessay, Laurent Durieux, and
  Alessandra~Rodrigues Gomes.
\newblock High spatial resolution land use and land cover mapping of the
  brazilian legal amazon in 2008 using landsat-5/tm and modis data.
\newblock \emph{Acta Amazonica}, 46\penalty0 (3):\penalty0 291--302, 2016.

\bibitem[Anderson et~al.(2002)Anderson, Cobb, Korpela, Lebofsky, and
  Werthimer]{SETI}
David~P Anderson, Jeff Cobb, Eric Korpela, Matt Lebofsky, and Dan Werthimer.
\newblock Seti@ home: an experiment in public-resource computing.
\newblock \emph{Communications of the ACM}, 45\penalty0 (11):\penalty0 56--61,
  2002.

\bibitem[Arcanjo et~al.(2016)Arcanjo, Luz, Fazenda, and Ramos]{ARCANJO}
Jeferson~S Arcanjo, Eduardo~FP Luz, {\'A}lvaro~L Fazenda, and Fernando~M Ramos.
\newblock Methods for evaluating volunteers’ contributions in a deforestation
  detection citizen science project.
\newblock \emph{Future Generation Computer Systems}, 56:\penalty0 550--557,
  2016.

\bibitem[Bonney et~al.(2016)Bonney, Cooper, and Ballard]{bonney2016theory}
Rick Bonney, Caren Cooper, and Heidi Ballard.
\newblock The theory and practice of citizen science: Launching a new journal.
\newblock \emph{Citizen Science: Theory and Practice}, 1\penalty0 (1), 2016.

\bibitem[Bradford and Israel(2004)]{BRADFORD2004}
Bianca~Marissa Bradford and Glenn~D Israel.
\newblock \emph{Evaluating volunteer motivation for sea turtle conservation in
  Florida}.
\newblock University of Florida Cooperative Extension Service, Institute of
  Food and~…, 2004.

\bibitem[Cohn(2008)]{COHN2008}
Jeffrey~P Cohn.
\newblock Citizen science: Can volunteers do real research?
\newblock \emph{BioScience}, 58\penalty0 (3):\penalty0 192--197, 2008.

\bibitem[Cooper et~al.(2010)Cooper, Khatib, Treuille, Barbero, Lee, Beenen,
  Leaver-Fay, Baker, Popovi{\'c}, et~al.]{FOLDIT}
Seth Cooper, Firas Khatib, Adrien Treuille, Janos Barbero, Jeehyung Lee,
  Michael Beenen, Andrew Leaver-Fay, David Baker, Zoran Popovi{\'c}, et~al.
\newblock Predicting protein structures with a multiplayer online game.
\newblock \emph{Nature}, 466\penalty0 (7307):\penalty0 756, 2010.

\bibitem[Coura and Borges-Pereira(2010)]{coura2010chagas}
Jos{\'e}~Rodrigues Coura and Jos{\'e} Borges-Pereira.
\newblock Chagas disease: 100 years after its discovery. {A} systemic review.
\newblock \emph{Acta tropica}, 115\penalty0 (1-2):\penalty0 5--13, 2010.

\bibitem[Cranshaw and Kittur(2011)]{POLYMATH}
Justin Cranshaw and Aniket Kittur.
\newblock The polymath project: lessons from a successful online collaboration
  in mathematics.
\newblock In \emph{Proceedings of the SIGCHI Conference on Human Factors in
  Computing Systems}, pages 1865--1874, 2011.

\bibitem[Curtis(2015)]{CURTIS}
Vickie Curtis.
\newblock Motivation to participate in an online citizen science game: A study
  of foldit.
\newblock \emph{Science Communication}, 37\penalty0 (6):\penalty0 723--746,
  2015.

\bibitem[Dallaqua et~al.(2019)Dallaqua, Fazenda, and Faria]{foresteyes2019}
F.B.J.R. Dallaqua, A.L. Fazenda, and F.A. Faria.
\newblock {ForestEyes} project: Can citizen scientists help rainforests?
\newblock In \emph{{IEEE} 15th {International} {Conference} on {eScience}},
  pages 18--27. {IEEE}, 9 2019.

\bibitem[Dallaqua et~al.(2018)Dallaqua, Faria, and Fazenda]{DALLAQUA}
Fernanda B. J.~R. Dallaqua, Fabio~Augusto Faria, and Alvaro~L Fazenda.
\newblock Active learning approaches for deforested area classification.
\newblock In \emph{2018 31st SIBGRAPI Conference on Graphics, Patterns and
  Images (SIBGRAPI)}, pages 48--55. IEEE, 2018.

\bibitem[de~Assis et~al.()de~Assis, Ferreira, Vinhas, Maurano, de~Almeida,
  Nascimento, de~Carvalho, Camargo, and Maciel]{deterrabrasilis}
Luiz Fernando Ferreira~Gomes de~Assis, Karine~Reis Ferreira, L{\'u}bia Vinhas,
  Luis Maurano, Cl{\'a}udio~Aparecido de~Almeida, Jether~Rodrigues Nascimento,
  Andr{\'e} Fernandes~Ara{\'u}jo de~Carvalho, Claudinei Camargo, and
  Adeline~Marinho Maciel.
\newblock Terrabrasilis: A spatial data infrastructure for disseminating
  deforestation data from brazil.

\bibitem[de~Souza~Jr et~al.(2009)de~Souza~Jr, Hayashi, and
  Ver{\'\i}ssimo]{IMAZON}
Carlos~M de~Souza~Jr, Sanae Hayashi, and Adalberto Ver{\'\i}ssimo.
\newblock Near real-time deforestation detection for enforcement of forest
  reserves in mato grosso.
\newblock In \emph{Proceedings of Land Governance in Support of the Millennium
  Development Goals: Responding to New Challenges, World Bank Conference,
  Washington, DC}, 2009.

\bibitem[Diniz et~al.(2015)Diniz, de~Almeida~Souza, Santos, Dias, da~Luz,
  de~Moraes, Maia, Gomes, da~Silva~Narvaes, Valeriano, et~al.]{DETER}
Cesar~Guerreiro Diniz, Arleson~Antonio de~Almeida~Souza, Diogo~Corr{\^e}a
  Santos, Mirian~Correa Dias, Nelton~Cavalcante da~Luz, Douglas Rafael~Vidal
  de~Moraes, Janaina~Sant’Ana Maia, Alessandra~Rodrigues Gomes, Igor
  da~Silva~Narvaes, Dalton~M Valeriano, et~al.
\newblock Deter-b: The new amazon near real-time deforestation detection
  system.
\newblock \emph{Ieee journal of selected topics in applied earth observations
  and remote sensing}, 8\penalty0 (7):\penalty0 3619--3628, 2015.

\bibitem[Fischer et~al.(2012)Fischer, Schwamb, Schawinski, Lintott, Brewer,
  Giguere, Lynn, Parrish, Sartori, Simpson, et~al.]{PLANETHUNTERS}
Debra~A Fischer, Megan~E Schwamb, Kevin Schawinski, Chris Lintott, John Brewer,
  Matt Giguere, Stuart Lynn, Michael Parrish, Thibault Sartori, Robert Simpson,
  et~al.
\newblock Planet hunters: the first two planet candidates identified by the
  public using the kepler public archive data.
\newblock \emph{Monthly Notices of the Royal Astronomical Society},
  419\penalty0 (4):\penalty0 2900--2911, 2012.

\bibitem[Fritz et~al.(2017)Fritz, Fonte, and See]{FRITZ}
Steffen Fritz, Cid{\'a}lia Fonte, and Linda See.
\newblock The role of citizen science in earth observation, 2017.

\bibitem[Fritz et~al.(2012)]{fritz2012geo}
Steffen Fritz et~al.
\newblock Geo-{W}iki: {A}n online platform for improving global land cover.
\newblock \emph{Environmental Modelling \& Software}, 31:\penalty0 110--123,
  2012.

\bibitem[Gomes et~al.(2014)Gomes, Diniz, and Almeida]{gomes2014amazon}
Alessandra~R Gomes, C{\'e}sar~G Diniz, and Cl{\'a}udio~A Almeida.
\newblock Amazon regional center (inpe/cra) actions for brazilian amazon
  forest: Terraclass and capacity building projects.
\newblock \emph{Interdisciplinary Analysis and Modeling of Carbon-Optimized
  Land Management Strategies for Southern Amazonia}, page 101, 2014.

\bibitem[Grey(2009)]{GREY2009}
Fran{\c{c}}ois Grey.
\newblock Viewpoint: The age of citizen cyberscience.
\newblock \emph{Cern Courier}, 29, 2009.

\bibitem[Guerrini et~al.(2018)Guerrini, Majumder, Lewellyn, and
  McGuire]{guerrini2018citizen}
Christi~J Guerrini, Mary~A Majumder, Meaganne~J Lewellyn, and Amy~L McGuire.
\newblock Citizen science, public policy.
\newblock \emph{Science}, 361\penalty0 (6398):\penalty0 134--136, 2018.

\bibitem[Guide(2017)]{guide2017landsat}
Product Guide.
\newblock Landsat surface reflectance-derived spectral indices; 3.6 version.
\newblock \emph{Department of the Interior US Geological Survey (USGS): Reston,
  VA, USA}, 2017.

\bibitem[Gura(2013)]{gura2013citizen}
Trisha Gura.
\newblock Citizen science: amateur experts.
\newblock \emph{Nature}, 496\penalty0 (7444):\penalty0 259--261, 2013.

\bibitem[Haklay(2013)]{HAKLAY2013}
Muki Haklay.
\newblock Citizen science and volunteered geographic information: Overview and
  typology of participation.
\newblock In \emph{Crowdsourcing geographic knowledge}, pages 105--122.
  Springer, 2013.

\bibitem[Hansen et~al.(2008)Hansen, Shimabukuro, Potapov, and Pittman]{PRODES}
Matthew~C Hansen, Yosio~E Shimabukuro, Peter Potapov, and Kyle Pittman.
\newblock Comparing annual modis and prodes forest cover change data for
  advancing monitoring of brazilian forest cover.
\newblock \emph{Remote Sensing of Environment}, 112\penalty0 (10):\penalty0
  3784--3793, 2008.

\bibitem[Hansen et~al.(2013)Hansen, Potapov, Moore, Hancher, Turubanova,
  Tyukavina, Thau, Stehman, Goetz, Loveland, et~al.]{Hansen850}
Matthew~C Hansen, Peter~V Potapov, Rebecca Moore, Matt Hancher, SAA Turubanova,
  Alexandra Tyukavina, David Thau, SV~Stehman, SJ~Goetz, Thomas~R Loveland,
  et~al.
\newblock High-resolution global maps of 21st-century forest cover change.
\newblock \emph{science}, 342\penalty0 (6160):\penalty0 850--853, 2013.

\bibitem[Hansen et~al.(2016)Hansen, Krylov, Tyukavina, Potapov, Turubanova,
  Zutta, Ifo, Margono, Stolle, and Moore]{GLAD}
Matthew~C Hansen, Alexander Krylov, Alexandra Tyukavina, Peter~V Potapov,
  Svetlana Turubanova, Bryan Zutta, Suspense Ifo, Belinda Margono, Fred Stolle,
  and Rebecca Moore.
\newblock Humid tropical forest disturbance alerts using landsat data.
\newblock \emph{Environmental Research Letters}, 11\penalty0 (3):\penalty0
  034008, 2016.

\bibitem[Haralick et~al.(1973)Haralick, Shanmugam, et~al.]{HARALICK}
Robert~M Haralick, Karthikeyan Shanmugam, et~al.
\newblock Textural features for image classification.
\newblock \emph{IEEE Transactions on systems, man, and cybernetics}, \penalty0
  (6):\penalty0 610--621, 1973.

\bibitem[Haykin(1994)]{ANN}
Simon Haykin.
\newblock \emph{Neural networks: a comprehensive foundation}.
\newblock Prentice Hall PTR, 1994.

\bibitem[Hochachka et~al.(2012)Hochachka, Fink, Hutchinson, Sheldon, Wong, and
  Kelling]{HOCHACHKA}
Wesley~M Hochachka, Daniel Fink, Rebecca~A Hutchinson, Daniel Sheldon,
  Weng-Keen Wong, and Steve Kelling.
\newblock Data-intensive science applied to broad-scale citizen science.
\newblock \emph{Trends in ecology \& evolution}, 27\penalty0 (2):\penalty0
  130--137, 2012.

\bibitem[Holohan and Garg(2005)]{HOLOHAN}
Anne Holohan and Anurag Garg.
\newblock Collaboration online: The example of distributed computing.
\newblock \emph{Journal of Computer-Mediated Communication}, 10\penalty0
  (4):\penalty0 JCMC10415, 2005.

\bibitem[INPE(2020)]{prodes2019}
INPE.
\newblock A taxa consolidada de desmatamento por corte raso para os nove
  estados da {A}mazônia {L}egal ({AC}, {AM}, {AP}, {MA}, {MT}, {PA}, {RO},
  {RR} e {TO}) em $2019$ é de $10.129 km^2$.
\newblock \url{http://www.inpe.br/noticias/noticia.php?Cod_Noticia=5465}, 2020.
\newblock Accessed: 2021-02-13.

\bibitem[Irving(2016)]{irving2016maskslic}
Benjamin Irving.
\newblock maskslic: regional superpixel generation with application to local
  pathology characterisation in medical images.
\newblock \emph{arXiv preprint arXiv:1606.09518}, 2016.

\bibitem[Irwin(2018)]{science2018}
Aisling Irwin.
\newblock No phds needed: how citizen science is transforming research.
\newblock \emph{Nature}, 562\penalty0 (7728):\penalty0 480—482, October 2018.
\newblock ISSN 0028-0836.
\newblock \doi{10.1038/d41586-018-07106-5}.
\newblock URL \url{https://doi.org/10.1038/d41586-018-07106-5}.

\bibitem[Justice et~al.(1998)Justice, Vermote, Townshend, Defries, Roy, Hall,
  Salomonson, Privette, Riggs, Strahler, et~al.]{MODIS}
Christopher~O Justice, Eric Vermote, John~RG Townshend, Ruth Defries, David~P
  Roy, Dorothy~K Hall, Vincent~V Salomonson, Jeffrey~L Privette, George Riggs,
  Alan Strahler, et~al.
\newblock The moderate resolution imaging spectroradiometer (modis): Land
  remote sensing for global change research.
\newblock \emph{IEEE transactions on geoscience and remote sensing},
  36\penalty0 (4):\penalty0 1228--1249, 1998.

\bibitem[King and Lynch(1998)]{KING1998}
Karen~N King and Cynthia~V Lynch.
\newblock The motivation of volunteers in the nature conservancy-ohio chapter,
  a non-profit environmental organization.
\newblock \emph{Journal of Volunteer Administration}, 16\penalty0 (2):\penalty0
  5--11, 1998.

\bibitem[Kodinariya and Makwana(2013)]{kodinariya2013review}
Trupti~M Kodinariya and Prashant~R Makwana.
\newblock Review on determining number of cluster in k-means clustering.
\newblock \emph{International Journal}, 1\penalty0 (6):\penalty0 90--95, 2013.

\bibitem[Kosmala et~al.(2016)Kosmala, Wiggins, Swanson, and Simmons]{KOSMALA}
Margaret Kosmala, Andrea Wiggins, Alexandra Swanson, and Brooke Simmons.
\newblock Assessing data quality in citizen science.
\newblock \emph{Frontiers in Ecology and the Environment}, 14\penalty0
  (10):\penalty0 551--560, 2016.

\bibitem[Lauer et~al.(1997)Lauer, Morain, and Salomonson]{lauer1997landsat}
Donald~T Lauer, Stanley~A Morain, and Vincent~V Salomonson.
\newblock The landsat program: Its origins, evolution, and impacts.
\newblock \emph{Photogrammetric Engineering and Remote Sensing}, 63\penalty0
  (7):\penalty0 831--838, 1997.

\bibitem[Lellis et~al.(2019)Lellis, Nakayama, and Porfiri]{De_Lellis_2019}
Pietro~De Lellis, Shinnosuke Nakayama, and Maurizio Porfiri.
\newblock Using demographics toward efficient data classification in citizen
  science: a bayesian approach.
\newblock \emph{{PeerJ} Computer Science}, 5:\penalty0 e239, nov 2019.
\newblock \doi{10.7717/peerj-cs.239}.
\newblock URL \url{https://doi.org/10.7717%2Fpeerj-cs.239}.

\bibitem[Lin(1991)]{SHANNON1}
Jianhua Lin.
\newblock Divergence measures based on the shannon entropy.
\newblock \emph{IEEE Transactions on Information theory}, 37\penalty0
  (1):\penalty0 145--151, 1991.

\bibitem[Lintott et~al.(2008)Lintott, Schawinski, Slosar, Land, Bamford,
  Thomas, Raddick, Nichol, Szalay, Andreescu, et~al.]{LINTOTT}
Chris~J Lintott, Kevin Schawinski, An{\v{z}}e Slosar, Kate Land, Steven
  Bamford, Daniel Thomas, M~Jordan Raddick, Robert~C Nichol, Alex Szalay, Dan
  Andreescu, et~al.
\newblock Galaxy zoo: morphologies derived from visual inspection of galaxies
  from the sloan digital sky survey.
\newblock \emph{Monthly Notices of the Royal Astronomical Society},
  389\penalty0 (3):\penalty0 1179--1189, 2008.

\bibitem[Lloyd(1982)]{lloyd1982least}
Stuart Lloyd.
\newblock Least squares quantization in pcm.
\newblock \emph{IEEE transactions on information theory}, 28\penalty0
  (2):\penalty0 129--137, 1982.

\bibitem[Lombra{\~n}a~Gonz{\'a}lez et~al.(2012)Lombra{\~n}a~Gonz{\'a}lez,
  Harutyunyan, Segal, Zacharov, McIntosh, Jones, Giovannozzi, Rivkin, Marquina,
  Skands, et~al.]{LHC2012}
D~Lombra{\~n}a~Gonz{\'a}lez, A~Harutyunyan, B~Segal, I~Zacharov, E~McIntosh,
  PL~Jones, M~Giovannozzi, L~Rivkin, MA~Marquina, P~Skands, et~al.
\newblock Lhc@ home: a volunteer computing system for massive numerical
  simulations of beam dynamics and high energy physics events.
\newblock In \emph{Conf. Proc.}, volume 1205201, pages 505--507, 2012.

\bibitem[Long et~al.(2015)Long, Shelhamer, and Darrell]{long2015fully}
Jonathan Long, Evan Shelhamer, and Trevor Darrell.
\newblock Fully convolutional networks for semantic segmentation.
\newblock In \emph{Proceedings of the IEEE conference on computer vision and
  pattern recognition}, pages 3431--3440, 2015.

\bibitem[Lovejoy and Nobre(2018)]{amazonscience2018}
Thomas~E. Lovejoy and Carlos Nobre.
\newblock Amazon tipping point.
\newblock \emph{Science Advances}, 4\penalty0 (2), 2018.
\newblock \doi{10.1126/sciadv.aat2340}.
\newblock URL \url{https://advances.sciencemag.org/content/4/2/eaat2340}.

\bibitem[Luz et~al.(2014)Luz, Correa, Gonz{\'a}lez, Grey, and Ramos]{AFW14}
Eduardo~FP Luz, Felipe~RS Correa, Daniel~L Gonz{\'a}lez, Fran{\c{c}}ois Grey,
  and Fernando~M Ramos.
\newblock The forestwatchers: a citizen cyberscience project for deforestation
  monitoring in the tropics.
\newblock \emph{Human Computation}, 1\penalty0 (2):\penalty0 137--145, 2014.

\bibitem[MacDicken et~al.(2016)MacDicken, Jonsson, Pi{\~n}a, Maulo, Contessa,
  Adikari, Garzuglia, Lindquist, Reams, and D’Annunzio]{FAO}
K~MacDicken, {\"O}~Jonsson, L~Pi{\~n}a, S~Maulo, V~Contessa, Y~Adikari,
  M~Garzuglia, E~Lindquist, G~Reams, and R~D’Annunzio.
\newblock Global forest resources assessment 2015: how are the world's forests
  changing?
\newblock 2016.

\bibitem[Maciel and Vinhas(2019)]{macielreasoning}
Adeline~Marinho Maciel and Lubia Vinhas.
\newblock Reasoning about deforestation trajectories in {P}ar\'a state, brazil.
\newblock In \emph{Anais do XIX Simpósio Brasileiro de Sensoriamento Remoto},
  2019.

\bibitem[Maisonneuve et~al.(2009)Maisonneuve, Stevens, Niessen, and
  Steels]{NOISETUBE}
Nicolas Maisonneuve, Matthias Stevens, Maria~E Niessen, and Luc Steels.
\newblock Noisetube: Measuring and mapping noise pollution with mobile phones.
\newblock In \emph{Information technologies in environmental engineering},
  pages 215--228. Springer, 2009.

\bibitem[Marshall et~al.(2012)Marshall, Kleine, and Dean]{CORALWATCH}
N~Justin Marshall, Diana~A Kleine, and Angela~J Dean.
\newblock Coralwatch: education, monitoring, and sustainability through citizen
  science.
\newblock \emph{Frontiers in Ecology and the Environment}, 10\penalty0
  (6):\penalty0 332--334, 2012.

\bibitem[Martin(2015)]{martin2015edge}
Claude Martin.
\newblock \emph{On the {E}dge: {T}he {S}tate and {F}ate of the {W}orld's
  {T}ropical {R}ainforests}.
\newblock Greystone Books Ltd, 2015.

\bibitem[Neves et~al.(2017)Neves, Korting, Fonseca, de~Queiroz, Vinhas,
  Ferreira, and Escada]{neves2017terraclass}
Alana~Kasahara Neves, Thales~Sehn Korting, Leila Maria~Garcia Fonseca,
  Gilberto~Ribeiro de~Queiroz, L{\'u}bia Vinhas, Karine~Reis Ferreira, and
  Maria Isabel~Sobral Escada.
\newblock Terraclass x mapbiomas: Comparative assessment of legend and mapping
  agreement analysis.
\newblock In \emph{GEOINFO}, pages 295--300, 2017.

\bibitem[Ortega~Adarme et~al.(2020)Ortega~Adarme, Queiroz~Feitosa, Nigri~Happ,
  Aparecido De~Almeida, and Rodrigues~Gomes]{rs12060910}
Mabel Ortega~Adarme, Raul Queiroz~Feitosa, Patrick Nigri~Happ, Claudio
  Aparecido De~Almeida, and Alessandra Rodrigues~Gomes.
\newblock Evaluation of deep learning techniques for deforestation detection in
  the brazilian amazon and cerrado biomes from remote sensing imagery.
\newblock \emph{Remote Sensing}, 12\penalty0 (6), 2020.
\newblock ISSN 2072-4292.
\newblock \doi{10.3390/rs12060910}.
\newblock URL \url{https://www.mdpi.com/2072-4292/12/6/910}.

\bibitem[Petersen et~al.(2017)Petersen, Pintea, and Bourgault]{forestwatcher}
Rachael Petersen, Lilian Pintea, and Liz Bourgault.
\newblock Forest {W}atcher {B}rings {D}ata {S}traight to {E}nvironmental
  {D}efenders.
\newblock
  {http://blog.globalforestwatch.org/people/forest-watcher-brings-data-straight-to-environmental-defenders/},
  2017.
\newblock Accessed: 02-15-2020.

\bibitem[Raddick et~al.(2009)Raddick, Bracey, Gay, Lintott, Murray, Schawinski,
  Szalay, and Vandenberg]{RADDICK2009}
M~Jordan Raddick, Georgia Bracey, Pamela~L Gay, Chris~J Lintott, Phil Murray,
  Kevin Schawinski, Alexander~S Szalay, and Jan Vandenberg.
\newblock Galaxy zoo: Exploring the motivations of citizen science volunteers.
\newblock \emph{arXiv preprint arXiv:0909.2925}, 2009.

\bibitem[Schepaschenko et~al.(2019)]{schepaschenko2019recent}
Dmitry Schepaschenko et~al.
\newblock Recent advances in forest observation with visual interpretation of
  very high-resolution imagery.
\newblock \emph{Surveys in Geophysics}, 40\penalty0 (4):\penalty0 839--862,
  2019.

\bibitem[Silvertown(2009)]{SILVERTOWN2009}
Jonathan Silvertown.
\newblock A new dawn for citizen science.
\newblock \emph{Trends in ecology \& evolution}, 24\penalty0 (9):\penalty0
  467--471, 2009.

\bibitem[Smith et~al.(2013)Smith, Lynn, and Lintott]{ZOONIVERSE2013}
Arfon~M Smith, Stuart Lynn, and Chris~J Lintott.
\newblock An introduction to the zooniverse.
\newblock In \emph{First AAAI conference on human computation and
  crowdsourcing}, 2013.

\bibitem[Soares et~al.(2010)Soares, Santos, Vijaykumar, and Dutra]{SOARES2010}
Marinalva~Dias Soares, Rafael Santos, Nandamudi Vijaykumar, and Luciano Dutra.
\newblock Citizen science-based labeling of imprecisely segmented images: Case
  study and preliminary results.
\newblock In \emph{2010 Brazilian Symposium on Collaborative Systems-Simposio
  Brasileiro de Sistemas Colaborativos}, pages 87--94. IEEE, 2010.

\bibitem[Souza et~al.(2019)Souza, Vieira~Monteiro, Daleles~Rennó, Almeida,
  de~Morisson~Valeriano, Morelli, Vinhas, P.~Maurano, Adami, Sobral~Escada,
  da~Motta, and Amaral]{metodprodes2019}
Arlesson Souza, Antônio~Miguel Vieira~Monteiro, Camilo Daleles~Rennó,
  Cláudio~Aparecido Almeida, Dalton de~Morisson~Valeriano, Fabiano Morelli,
  Lubia Vinhas, Luis~Eduardo P.~Maurano, Marcos Adami, Maria~Isabel
  Sobral~Escada, Marisa da~Motta, and Silvana Amaral.
\newblock Metodologia utilizada nos projetos {P}{R}{O}{D}{E}{S} e
  {D}{E}{T}{E}{R}.
\newblock \emph{S{\~a}o Jos{\'e} dos Campos: INPE}, 2019.

\bibitem[Souza and Azevedo(2017)]{souza2017mapbiomas}
C~Souza and Tasso Azevedo.
\newblock Mapbiomas general handbook.
\newblock \emph{MapBiomas: S{\~a}o Paulo, Brazil}, pages 1--23, 2017.

\bibitem[Stainforth et~al.(2002)Stainforth, Kettleborough, Martin, Simpson,
  Gillis, Akkas, Gault, Collins, Gavaghan, and Allen]{CLIMATE2002}
Dave Stainforth, Jamie Kettleborough, Andrew~P Martin, Andrew Simpson, Richard
  Gillis, Ali Akkas, Richard Gault, Mat Collins, David Gavaghan, and Myles
  Allen.
\newblock Climateprediction. net: Design principles for publicresource modeling
  research.
\newblock In \emph{IASTED PDCS}, pages 32--38. Citeseer, 2002.

\bibitem[Sullivan et~al.(2009)Sullivan, Wood, Iliff, Bonney, Fink, and
  Kelling]{EBIRD2009}
Brian~L Sullivan, Christopher~L Wood, Marshall~J Iliff, Rick~E Bonney, Daniel
  Fink, and Steve Kelling.
\newblock ebird: A citizen-based bird observation network in the biological
  sciences.
\newblock \emph{Biological Conservation}, 142\penalty0 (10):\penalty0
  2282--2292, 2009.

\bibitem[Sullivan et~al.(2014)Sullivan, Aycrigg, Barry, Bonney, Bruns, Cooper,
  Damoulas, Dhondt, Dietterich, Farnsworth, et~al.]{EBIRD2014}
Brian~L Sullivan, Jocelyn~L Aycrigg, Jessie~H Barry, Rick~E Bonney, Nicholas
  Bruns, Caren~B Cooper, Theo Damoulas, Andr{\'e}~A Dhondt, Tom Dietterich,
  Andrew Farnsworth, et~al.
\newblock The ebird enterprise: an integrated approach to development and
  application of citizen science.
\newblock \emph{Biological Conservation}, 169:\penalty0 31--40, 2014.

\bibitem[Tuia et~al.(2011)Tuia, Volpi, Copa, Kanevski, and
  Munoz-Mari]{TUIA2011}
Devis Tuia, Michele Volpi, Loris Copa, Mikhail Kanevski, and Jordi Munoz-Mari.
\newblock A survey of active learning algorithms for supervised remote sensing
  image classification.
\newblock \emph{IEEE Journal of Selected Topics in Signal Processing},
  5\penalty0 (3):\penalty0 606--617, 2011.

\bibitem[Walker et~al.(2010)Walker, Stickler, Kellndorfer, Kirsch, and
  Nepstad]{PRODES3}
Wayne~S Walker, Claudia~M Stickler, Josef~M Kellndorfer, Katie~M Kirsch, and
  Daniel~C Nepstad.
\newblock Large-area classification and mapping of forest and land cover in the
  brazilian amazon: A comparative analysis of alos/palsar and landsat data
  sources.
\newblock \emph{IEEE Journal of selected topics in applied earth observations
  and remote sensing}, 3\penalty0 (4):\penalty0 594--604, 2010.

\bibitem[Westphal et~al.(2005)Westphal, Butterworth, Snead, Craig, Anderson,
  Jones, Brownlee, Farnsworth, and Zolensky]{STARDUST}
Andrew~J Westphal, Anna~L Butterworth, Christopher~J Snead, Nahide Craig, David
  Anderson, Steven~M Jones, Donald~E Brownlee, Richard Farnsworth, and
  Michael~E Zolensky.
\newblock Stardust@ home: a massively distributed public search for
  interstellar dust in the stardust interstellar dust collector.
\newblock 2005.

\end{thebibliography}

\end{document}